\documentclass[11pt,a4paper]{article}
\usepackage[hyperref]{acl2021}
\usepackage{times}
\usepackage{latexsym}

\usepackage{microtype}

\usepackage[utf8]{inputenc}  
\usepackage[english]{babel}                   % English

\usepackage{pdfcomment}                       % alt text on images

\usepackage{graphicx}
\usepackage{subcaption}

\usepackage{algorithm}
\usepackage{algorithmic}
\usepackage{listings}                         % Source code blocks

\usepackage{booktabs}
\usepackage{multirow, tabularx}               % Tables
\usepackage{tabulary}                         % Table wrap text

\usepackage{color}  
\usepackage{footnote}                         % Footnotes
\usepackage[nolist,nohyperlinks]{acronym}     % Acronyms

\usepackage{amssymb}                          % Symbols (e.g., checkmark)
\usepackage{amsopn}                           % New math operators

\usepackage{colortbl}                       % Colors on table

\usepackage{gb4e}                             % Linguistics examples
\noautomath

\newcommand{\red}[1]{} 
\renewcommand{\red}[1]{{\color{red} {#1}}}

\newcommand{\blue}[1]{} 
\renewcommand{\blue}[1]{{\color[rgb]{0,0,0.75} {#1}}}

\graphicspath{{./img/}}

\lstdefinestyle{plain}{
  belowcaptionskip=1\baselineskip,
  aboveskip=5pt,
  belowskip=5pt,
  breaklines=true,
  frame=lines,
  xleftmargin=\parindent,
  showstringspaces=false,
  basicstyle=\ttfamily,
}

\newcommand{\STAB}[1]{\begin{tabular}{@{}c@{}}#1\end{tabular}}

\aclfinalcopy % Uncomment this line for the final submission

\def\aclpaperid{2700} %  Enter the acl Paper ID here

\title{{O}nline {L}earning Meets {M}achine {T}ranslation Evaluation: Finding the Best Systems with the Least Human Effort}

\author{Vânia Mendonça$^{1,2}$,  Ricardo Rei$^{1,2,3}$, 
 Luísa Coheur$^{1,2}$,\\
{\bf Alberto Sardinha$^{1,2}$, Ana Lúcia Santos$^{4,5}$}\\
$^1$ INESC-ID Lisboa, Portugal \\
$^2$ Instituto Superior T\'{e}cnico, Universidade de Lisboa, Portugal \\
$^3$ Unbabel AI, Lisboa, Portugal\\
$^4$ Centro de Linguística da Universidade de Lisboa, Portugal \\
$^5$ Faculdade de Letras da Universidade de Lisboa, Portugal \\
 \small \texttt{\{vania.mendonca, luisa.coheur,  jose.alberto.sardinha\}@tecnico.ulisboa.pt},\\  
 \small \texttt{ricardo.rei@unbabel.com, als@letras.ulisboa.pt} \\
  }

\date{}

\begin{document}
\maketitle

%%%%%%%%%%%%%%%%%%%%%%%%%%%%%%%%%%%%%%%%%%%%%%%%%%%%%%%%%%%%%%%%%%%%%%%%%%%%%%%%
% ACRONYM

\begin{acronym}[EEGPET]

    \acrodefplural{RQ}[RQ]{research questions}
    
    \acro{CRF}      {Conditional Random Fields}
    \acro{EWAF}     {Exponentially Weighted Average Forecaster}
    \acro{EXP4}     {Exponential-weighting for Exploration and Exploitation with Experts}
    \acro{EXP3}     {Exponential-weighting for Exploration and Exploitation}
    \acro{$F_1$}    {F-Score}
    \acro{FN}       {false negatives}
    \acro{FP}       {false positives}
    \acro{LDC}      {Linguistic Data Consortium}
    \acro{HMM}      {Hidden Markov Model}
    \acro{MDP}      {Markov Decision Problem}
    \acro{MT}       {Machine Translation}
    \acro{MQM}      {Multidimensional Quality Metric}
    \acro{NER}      {Named Entity Recognition}
    \acro{NLP}      {Natural Language Processing}
    \acro{NLU}      {Natural Language Understanding}
    \acro{POMDP}    {Partially Observable Markov Decision Problem}
    \acro{RQ}       {research question}
    \acro{StDev}    {Standard Deviation}
    \acro{TN}       {true negatives}
    \acro{TP}       {true positives}
    \acro{UCB}      {Upper Confidence Bound}
    \acro{WM}       {Weighted Majority}
    \acro{WMT}      {Conference on Machine Translation}

\end{acronym}

%%%%%%%%%%%%%%%%%%%%%%%%%%%%%%%%%%%%%%%%%%%%%%%%%%%%%%%%%%%%%%%%%%%%%%%%%%%%%%%%
% ABSTRACT
\begin{abstract}

In \acl{MT}, assessing the quality of a large amount of automatic translations can be challenging. Automatic metrics are not reliable when it comes to high performing systems. In addition, resorting to human evaluators can be expensive, especially when evaluating multiple systems. 
To overcome the latter challenge, we propose a novel application of online learning that, given an ensemble of \acl{MT} systems, dynamically converges to the best systems, by taking advantage of the human feedback available. 
Our experiments on \acs{WMT}'19 datasets show that our online approach quickly converges to the top-3 ranked systems for the language pairs considered, despite the lack of human feedback for many translations.

\end{abstract}

%%%%%%%%%%%%%%%%%%%%%%%%%%%%%%%%%%%%%%%%%%%%%%%%%%%%%%%%%%%%%%%%%%%%%%%%%%%%%%%%
% INTRO
\section{Introduction}
\label{sec:intro}

In \ac{MT}, measuring the quality of a large amount of automatic translations can be a challenge. Automatic metrics like {\sc Bleu} \citep{papineni-bleu} remain popular due to their fast and free computations. Yet, in the last few years we have seen that, as \ac{MT} quality improves, automatic metrics become less reliable \citep{ma-etal-2019-results, mathur-etal-2020-tangled}. For example, in the \ac{WMT}'19 News Translation shared task, the winning system according to human annotators was not even in the top-5 according to {\sc Bleu} \citep{Barrault2019}.
On the other hand, using human assessments can be expensive, especially when evaluating multiple systems. In a real world scenario, given an arbitrary number of \ac{MT} systems, one would need to evaluate them individually to find the best systems for a given language pair. However, that requires a considerable effort and there may not be enough human annotators to evaluate all the systems' translations. For instance, in the aforementioned \ac{WMT}'19 shared task, many translations from the competing systems did not receive any human assessment. 

Given an ensemble of competing, independent \ac{MT} systems, how can we dynamically find the best ones for a given language pair, while making the most of existing human feedback? 
To address this question, we present a novel application of online learning to \ac{MT}: each \ac{MT} system in the ensemble is assigned to a weight, and the systems' weights are updated considering human feedback regarding the quality of their translations at each iteration. 
We use online learning algorithms with theoretical performance guarantees, under the frameworks of prediction with expert advice \citep{cesa-bianchi06} and multi-armed bandits \citep{Herbert1952,Lai1985}. 

We contribute with an online \ac{MT} ensemble that allows to reduce human effort by immediately incorporating human feedback in order to dynamically converge to the best systems\footnote{The code for our experiments can be found in \url{https://github.com/vania-mendonca/MTOL}}. Our experiments on \ac{WMT}'19 News Translation test sets show that our online approaches indeed converge to the shared task's official top-3 systems (or to a subset of them) in just a few hundred iterations for all the language pairs experimented. Moreover, it does so while coping with the aforementioned lack of human assessments for many translations, through the use of fallback metrics. 

%%%%%%%%%%%%%%%%%%%%%%%%%%%%%%%%%%%%%%%%%%%%%%%%%%%%%%%%%%%%%%%%%%%%%%%%%%%%%%%%
% EXPERT ADVICE
\section{Online learning frameworks}
\label{sec:experts}

To provide some background on our proposal, we start by describing the online learning frameworks that we apply in this paper: prediction with expert advice and multi-armed bandits.

A problem of prediction with expert advice can be described as an iterative game between a {\em forecaster} and the {\em environment}, in which the forecaster seeks advice from different sources ({\em experts}) in order to provide the best forecast \citep{cesa-bianchi06}. 
At each iteration $t$, the forecaster consults the predictions $\hat{p}_{j,t}, j = 1 \ldots J,$ made by a set of $J$ weighted experts, in the decision space $\mathcal{D}$. 
Considering these predictions, the forecaster makes its own prediction, $\hat{p}_{f,t}\in\mathcal{D}$. At the same time, the environment reveals an outcome $y_t$ in the decision space $\mathcal{Y}$ (which may not necessarily be the same as $\mathcal{D}$). 

A well-established algorithm to learn the experts' weights in this framework is \ac{EWAF} \citep{cesa-bianchi06}. 
In \ac{EWAF}, the prediction made by the forecaster is randomly selected following the probability distribution based on the experts' weights $\omega_{1,t-1} \ldots \omega_{J,t-1}$:%, as shown in Eq.~\ref{eq:PEWAF}:

\begin{equation}
\hat{p}_{f,t} = \frac{\sum_{j=1}^J\omega_{j,t-1} p_{j,t}}{\sum_{j=1}^J\omega_{j,t-1}}.
\label{eq:PEWAF}
\end{equation}

At the end of each iteration, the forecaster and each of the experts receive a non-negative loss based on the outcome $y_t$ revealed by the environment ($\ell_{f,t}$ and $\ell_{j,t}$, respectively). 
The weight $\omega_{j,t}$ of each expert $j = 1 \ldots J$ is then updated according to the loss received by each expert, as follows:

\begin{equation}
\omega_{j,t}=\omega_{j,t-1}e^{-\eta\ell_{j,t}}
\label{eq:weightUpdate}
\end{equation} 

If the parameter $\eta$ is set to $\sqrt{\frac{8\log{J}}{T}}$, 
it can be shown that the forecaster quickly converges to the performance of the best expert after $T$ iterations \citep{cesa-bianchi06}. 

Prediction with expert advice assumes that both the forecaster and all the experts receive a loss once the environment's outcome is revealed. 
However, this assumption may not always hold (i.e., there may not always be an environment's explicit feedback or a way to obtain the loss for all the experts). 
Thus, we consider a related class of problems, {\em multi-armed bandits}, in which the environment's outcome is unknown \citep{Herbert1952,Lai1985}. In this class of problems, one starts by attempting to estimate the means of the loss distributions for each expert (also known as {\em arm}) in the first iterations (the exploration phase), and when the forecaster has a high level of confidence in the estimated values, one may keep choosing the prediction with the smallest estimated loss (the exploitation phase).

A popular online algorithm for adversarial multi-armed bandits is \ac{EXP3} \citep{Auer1995}. 
At each iteration $t$, the forecaster's {\em action} is randomly selected according to the probability distribution given by the weights of each arm $j$: 

\begin{equation}
\hat{p}_{f,t} = \frac{\omega_j}{ \sum_{j'=1}^J \omega_{j'}}
\label{eq:predictionEXP3}
\end{equation} 

In this framework, the forecaster is only able to measure the loss of the action it selects at each iteration, but it cannot measure the loss of other possible actions. Thus, only the weight of the arm associated with this action is updated, as follows:

\begin{equation}
\omega_{j,t}=\omega_{j,t-1}e^{-\eta\hat{\ell}_{j,t}}
\label{eq:weightUpdateEXP3}
\end{equation} 
where $\hat{\ell}_{j,t} = \frac{\ell_{j,t}}{p_{j,t}}$ and $p_{j,t}$ is the probability of choosing arm $j$ at iteration $t$. 
By setting $\eta$ to $\sqrt{\frac{2 log J}{T |\mathcal{A}|}}$ (where $|\mathcal{A}|$ is the number the actions available, and may be the same as the number of arms $J$), it can be shown that the forecaster quickly converges to the performance of the best arm.

Both of these frameworks are relatively under-explored in \acs{NLP}, despite their potential to converge to the best performing approach available in scenarios where feedback is naturally present. Therefore, we propose to apply them in order to find the best \ac{MT} models with little human feedback.

%%%%%%%%%%%%%%%%%%%%%%%%%%%%%%%%%%%%%%%%%%%%%%%%%%%%%%%%%%%%%%%%%%%%%%%%%%%%%%%%
% EXP SETUP
\section{\acl{MT} with Online Learning}
\label{sec:onlineMT}

In this work, we consider the following scenario as the starting point: there is an ensemble composed of an arbitrary number of \ac{MT} systems; given a segment from a source language corpus, each system outputs a translation in the target language; then, the quality of the translations produced by each of the available systems is assessed by one or more human evaluators with a score reflecting their quality. 

We frame this scenario as an online learning problem under two different frameworks: (i) prediction with expert advice (using \ac{EWAF} as the learning algorithm), and (ii) multi-armed bandits (using \ac{EXP3} as the learning algorithm). The decision on whether to use one or another framework in an \ac{MT} scenario depends on whether there is human feedback available for the translations outputted by all the available systems or only for the final choice of the ensemble of systems. 

\begin{figure}[t]
\centering
\pdftooltip{
\includegraphics[width=1\columnwidth]{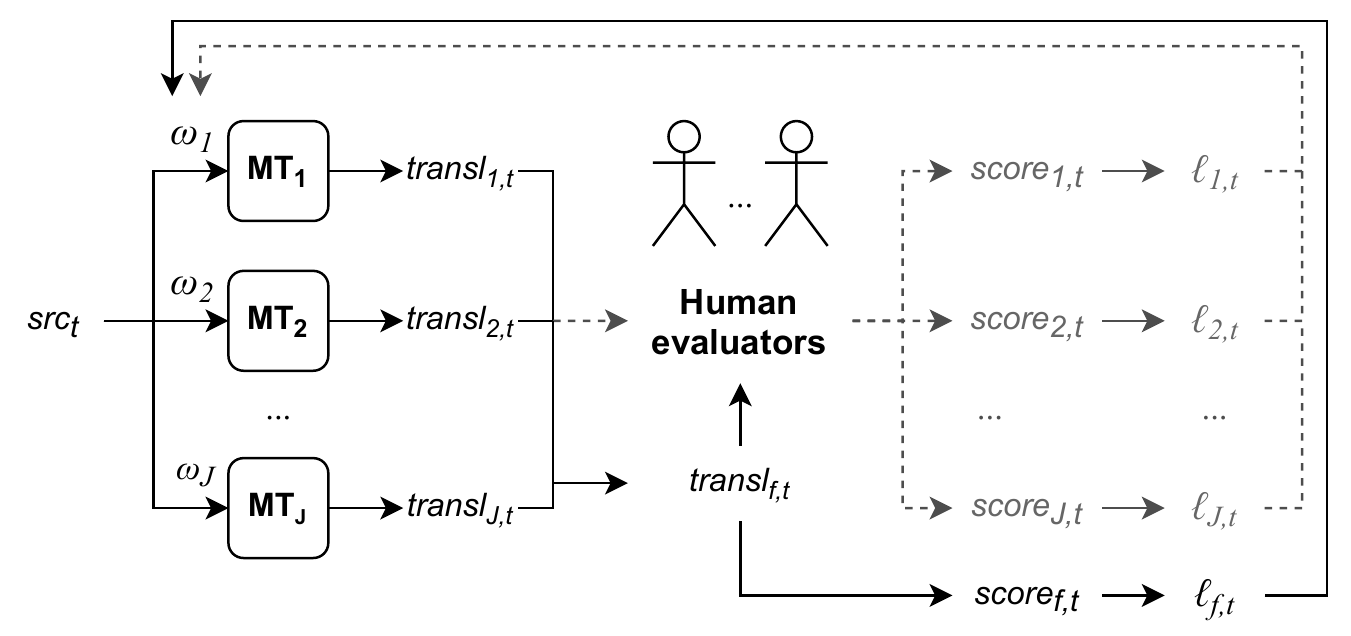}}
{Diagram displaying an arbitrary number of J machine translation systems, each assigned to a weight omega J. Each machine translation system outputs transl t j. They flow to the chosen translation transl t f, and also to the human evaluators (in this case, with dashed lines). The human translators output a score t j for each translation that leads to a loss l t j. This loss then flows back to the weights omega t j assigned to each machine translation system.}%
\caption{Overview of the online learning process applied to \ac{MT}, at each iteration $t$. The grey dashed arrows represent flows that only occur when using prediction with expert advice.}
\label{fig:MTOnline}
\end{figure}

An overview of the online learning process is shown in Fig.\ref{fig:MTOnline}, and can be summed up as follows. Each \ac{MT} system is an expert (or arm) $j = 1 \ldots J$, associated with a weight $\omega_j$ (all the systems start with same weights). At each iteration $t$, a segment $src_t$ is selected from the source language corpus and handed to all the \ac{MT} systems. Each system outputs a translation $transl_{j,t}$ in the target language, and one of these translations is selected as the forecaster's action according to the probability distribution given by the systems' weights (Eq.\ref{eq:PEWAF} for \ac{EWAF} and Eq.~\ref{eq:predictionEXP3} for \ac{EXP3}). 
The chosen translation $transl_{f,t}$ (when using \ac{EXP3}) or the translations outputted by all the systems (when using \ac{EWAF}) receive a human assessment score\footnote{If multiple human assessments were made for the same translation, $score_{j,t}$ is the average of the scores received.} $score_{j,t}$, from which the loss $\ell_{j,t}$ is derived for the respective \ac{MT} system. Finally, the weight of the chosen system or the weights of all the systems are updated as a function of the loss received, according to Eq.\ref{eq:weightUpdateEXP3} (when using \ac{EXP3}) and Eq.\ref{eq:weightUpdate}  (when using \ac{EWAF}), respectively (where $\ell_{j,t} = - score_{j,t}$). 

%%%%%%%%%%%%%%%%%%%%%%%%%%%%%%%%%%%%%%%%%%%%%%%%%%%%%%%%%%%%%%%%%%%%%%%%%%%%%%%%

\section{Experimental setup}
\label{sec:setup}

To validate our proposal, we designed an experiment using data from an \ac{MT} shared task. The main questions addressed by our experiment are: (i) whether an online learning approach can give a greater weight to the top performing systems for each language pair according to the shared task's official ranking, and (ii) if so, how quickly (i.e., how many translations need to be assessed by human evaluators in order to find the best system). 

Below we detail the datasets used (Section~\ref{sec:datasets}) and the feedback sources considered (Section~\ref{sec:feedback}), as well as other experimental decisions (Section~\ref{sec:design}).

\subsection{Datasets}
\label{sec:datasets}
We used the test datasets made available by the \ac{WMT}'19 News Translation shared task \citep{Barrault2019}. 
For each language pair, each source segment is associated with the following information: 

\begin{itemize}
    \item A reference translation in the target language (produced specifically for the task);
    \item The automatic translation outputted by each system competing in the task for that language pair;
    \item The average score obtained by each automatic translation, according to human assessments made by one or more human evaluators, in two formats: a raw score in [0;100] and a z-score in [$-\infty;+\infty$]. Not all the automatic translations received a human assessment;
    \item The number of human evaluators for each automatic translation (if there were any).
\end{itemize}

\begin{table*}[t]
\centering
\begin{tabular}{@{}rccccc@{}}
\toprule
\multicolumn{1}{l}{}       & \multicolumn{1}{c}{{\tt en-de}} & \multicolumn{1}{c}{{\tt fr-de}} & \multicolumn{1}{c}{{\tt de-cs}} & \multicolumn{1}{c}{{\tt gu-en}} & \multicolumn{1}{c}{{\tt lt-en}} \\ \midrule
Test set size (\# segments)              & 1997                      & 1701                      & 1997                      & 1016                      & 1000                      \\
Competing systems          & 22                        & 10                        & 11                        & 12                        & 11                        \\
Human assessments coverage & 86.80\%                   & 23.52\%                   & 62.94\%                   & 75.00\%                   & 100.00\%                  \\ \bottomrule
\end{tabular}
\caption{Overview of the language pairs considered in our experiments.}
\label{tab:testsets}
\end{table*}

\begin{table*}[!ht]
\centering
\begin{tabular}{@{}clrr@{}}
\toprule
\multicolumn{1}{l}{}            & \textbf{Top 3}      & \multicolumn{1}{l}{\textbf{z-score}} & \multicolumn{1}{l}{\textbf{Raw score}} \\ \midrule

\multirow{3}{*}{\STAB{\rotatebox[origin=c]{90}{\texttt{en-de}}}} 
                                & Facebook-FAIR \citep{Ng2019}       & 0.347                                & 90.3                                   \\
                                & Microsoft-sent-doc \citep{Junczys-Dowmunt2019}  & 0.311                                & 93.0                                   \\
                                & Microsoft-doc-level \citep{Junczys-Dowmunt2019} & 0.296                                & 92.6                                   \\ \cmidrule{1-4}
\multirow{3}{*}{\STAB{\rotatebox[origin=c]{90}{\texttt{fr-de}}}} 
                                & MSRA-MADL \citep{Xia2019}           & 0.267                                & 82.4                                   \\
                                & eTranslation \citep{Oravecz2019}       & 0.246                                & 81.5                                   \\
                                & LIUM \citep{Bougares2019}               & 0.082                                & 78.5                                   \\ \cmidrule{1-4}
\multirow{3}{*}{\STAB{\rotatebox[origin=c]{90}{\texttt{de-cs}}}} 
                                & online-Y            & 0.426                                & 63.9                                   \\
                                & online-B            & 0.386                                & 62.7                                   \\
                                & NICT \citep{Dabre2019}               & 0.367                                & 61.4                                   \\ \cmidrule{1-4}
\multirow{3}{*}{\STAB{\rotatebox[origin=c]{90}{\texttt{gu-en}}}} 
                                & NEU \citep{Li2019}                & 0.210                                & 64.8                                   \\
                                & UEDIN \citep{Bawden2019}              & 0.126                                & 61.7                                   \\
                                & GTCOM-Primary \citep{Bei2019}      & 0.100                                & 59.4                                   \\ \midrule
\multirow{3}{*}{\STAB{\rotatebox[origin=c]{90}{\texttt{lt-en}}}} 
                                & GTCOM-Primary \citep{Bei2019}      & 0.234                                & 77.4                                   \\
                                & tilde-nc-nmt \citep{Pinnis2019}       & 0.216                                & 77.5                                   \\
                                & NEU \citep{Li2019}                & 0.213                                & 77.0                                   \\ \bottomrule

\end{tabular}
\caption{Top 3 performing systems for each language pair in the \ac{WMT}'19 News Translation shared task \citep{Barrault2019}. The systems named ``online-[letter]'' correspond to publicly available translation services and were anonimized in the shared task.}
\label{tab:top3systems}
\end{table*}

For brevity, we focused on five language pairs, listed in Table~\ref{tab:testsets}. The official top 3 systems for each pair, according to the average z-score, are shown in Table~\ref{tab:top3systems}. Our choice of language pairs attempts to capture as many different phenomena as possible with the fewest pairs:

\begin{itemize}
    \item English $\rightarrow$ German ({\tt en-de}): This is the language pair with the most competitors and does not have a clear winning system (the winner differs depending on whether one considers the z-score or the raw score); 
    \item French $\rightarrow$ German ({\tt fr-de}): Unlike most language pairs, this pair features two languages other than English. Moreover, there is a strong imbalance between translations lacking human assessments and translations that received at least one assessment;
    \item German $\rightarrow$ Czech ({\tt de-cs}): Besides featuring two languages other than English, this pair stands out as it was devised as an unsupervised task (i.e., English was used as a ``hub'' language);
    \item Gujarati $\rightarrow$ English ({\tt gu-en}): This is one of the task's {\em low-resource} language pairs (i.e., whose test set is half the size of most language-pairs in the task), and is one where there may be more linguistic differences between the source and the target languages (e.g., different writing systems). Unlike {\tt en-de}, there is a clear winner considering both raw and z-score. Moreover, three of the competing systems did not receive any human assessment on their translations;
    \item Lithuanian $\rightarrow$ English ({\tt lt-en}): This is another {\em low-resource} language pair, with a rather competitive top 3. Unlike most language pairs, all the translations submitted by the competing systems for this pair received a human assessment.
\end{itemize}

For all these language pairs (except English $\rightarrow$ German), each segment was given an assessment score considering only the reference translation (and without access to the segment's context within the document to which it belongs). For English $\rightarrow$ German, scores were given considering the source segment instead of the reference, and evaluators had access to the segment's context within the document.

\subsection{Human feedback}
\label{sec:feedback}

\begin{figure}[!ht]
\centering
\pdftooltip{
\includegraphics[width=0.8\columnwidth]{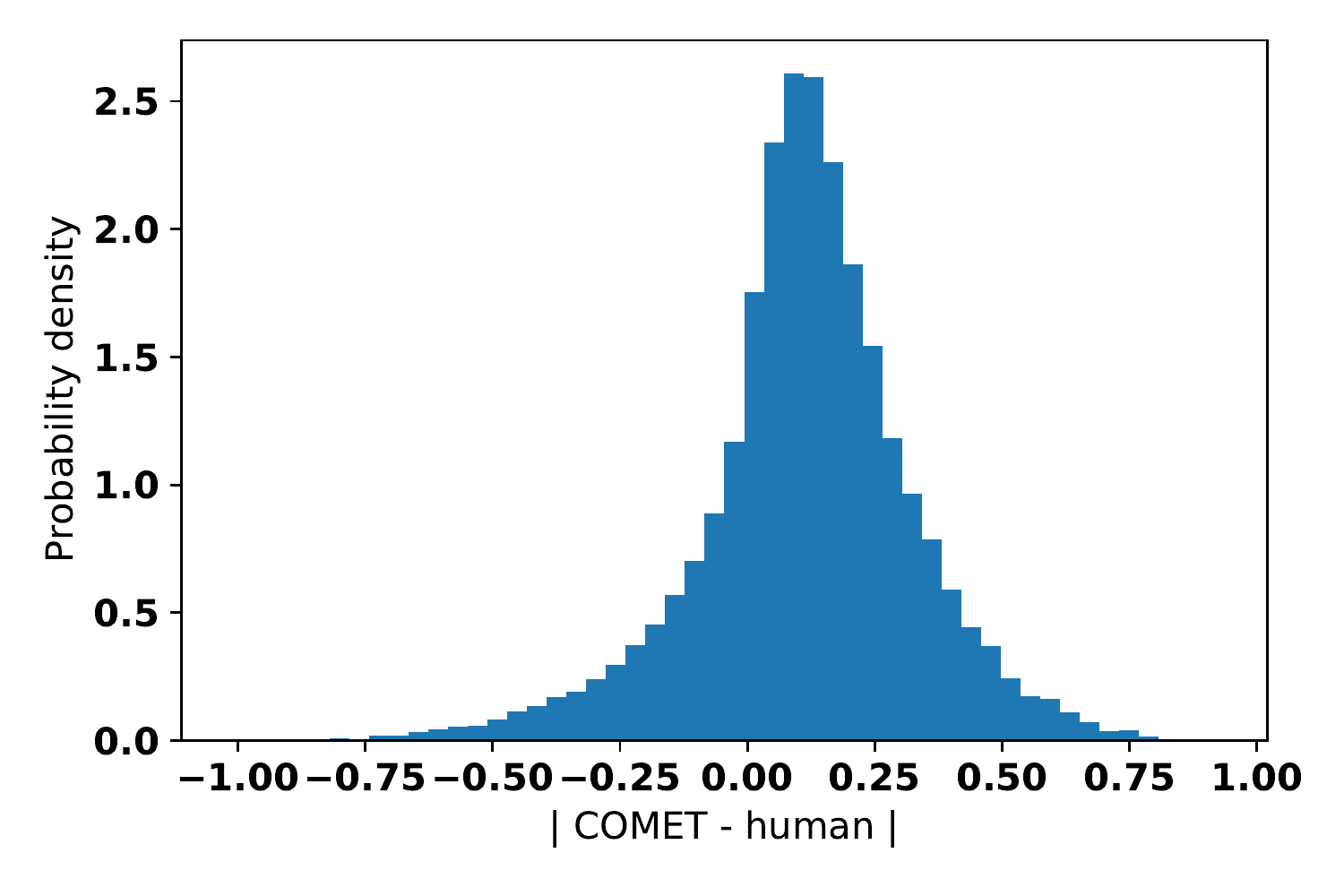}}
{Bar plot displaying the probability density of the difference between the Comet and human scores. the higher density values are located near a difference of zero.}
\caption{Distribution of the difference between the (existing) human assessments and {\sc Comet} scores for the same translations, on the {\tt en-de} test set.}
\label{fig:humancomet}
\end{figure}

A key condition for applying online learning to this scenario is the availability of feedback. We use the human assessment raw scores\footnote{Although we assume an absolute scale of scores in [0;100] in our experiments, our approach could be applied to any other level of granularity.} present in the test sets as a feedback source to compute the loss and update the weight of each \ac{MT} system, as already suggested in Section~\ref{sec:onlineMT}. 
However, not all translations received human assessments (recall Table~\ref{tab:testsets}). To cope with this issue, we designed different variants of this loss function, following different fallback strategies:

\begin{itemize}
    \item {\tt human-zero}: If there is no human assessment for the current translation, a score of zero is returned (leading to an unchanged weight on that iteration);
    \item {\tt human-avg}: If there is no human assessment for the current translation, the average of the previous scores received by the system behind that translation is returned as the current score;
    \item {\tt human-comet}: If there is no human assessment for the current translation, the {\sc Comet} score \citep{rei-etal-2020-comet} between the translation and the pair source/reference available in the corpus is returned as the current score. We pre-trained\footnote{We trained this metric from scratch following the hyper-parameters described in \citet{rei-etal-2020-unbabels}, except that we used the raw scores instead of the z-normalized scores.} this automatic metric on the datasets of previous shared tasks (\ac{WMT}'17 \citep{Bojar2017} and \ac{WMT}'18 \citep{Bojar2018}). Thus, for most translations, it displays a small difference regarding the existing human scores (see Fig.~\ref{fig:humancomet} for the case of {\tt en-de}). Moreover, this metric correlates better with ratings by professional translators than the \ac{WMT} scores \citep{Freitag2021}.
\end{itemize}

\subsection{Experimental design}
\label{sec:design}

For each language pair, we shuffled the test set once, so that the performance of the online algorithms would not be biased by the original order of the segments in the test set. We ran \ac{EWAF} once for each loss function, and we ran \ac{EXP3} 10 times per loss function and report the average weights obtained across runs, since \ac{EXP3}'s weight evolution is critically influenced by the random choice of an arm at each iteration. We normalized the translation scores $score_{j,t}$ to be in the interval $[0,1]$ and rounded them to two decimal places, to avoid exploding weight values due to the exponential update rule.

%%%%%%%%%%%%%%%%%%%%%%%%%%%%%%%%%%%%%%%%%%%%%%%%%%%%%%%%%%%%%%%%%%%%%%%%%%%%%%%%
% EXP RESULTS
\section{Results and discussion}
\label{sec:results}

\begin{table*}[t]
\centering
\begin{tabular}{@{}clrrrrrrrrrrrr@{}}
\toprule
\multicolumn{2}{r}{\textbf{Iteration}}                    & \multicolumn{2}{c}{10}                                      & \multicolumn{2}{c}{50}                                      & \multicolumn{2}{c}{100}                                     & \multicolumn{2}{c}{500}                                     & \multicolumn{2}{c}{1000}                                    & \multicolumn{2}{c}{1997}                                    \\\midrule
\multicolumn{2}{r}{\textbf{Top}}                          & \multicolumn{1}{c}{1}        & \multicolumn{1}{c}{3}        & \multicolumn{1}{c}{1}        & \multicolumn{1}{c}{3}        & \multicolumn{1}{c}{1}        & \multicolumn{1}{c}{3}        & \multicolumn{1}{c}{1}        & \multicolumn{1}{c}{3}        & \multicolumn{1}{c}{1}        & \multicolumn{1}{c}{3}        & \multicolumn{1}{c}{1}        & \multicolumn{1}{c}{3}        \\ \midrule

\multirow{3}{*}{\STAB{\rotatebox[origin=c]{90}{\small \acs{EWAF}}}} & human-zero                            & \cellcolor[HTML]{FFFFFF}0.00    & \cellcolor[HTML]{E7F2E2}0.33 & \cellcolor[HTML]{FFFFFF}0.00    & \cellcolor[HTML]{FFFFFF}0.00    & \cellcolor[HTML]{FFFFFF}0.00    & \cellcolor[HTML]{E7F2E2}0.33 & \cellcolor[HTML]{B6D7A8}1.00 & \cellcolor[HTML]{CFE5C5}0.67 & \cellcolor[HTML]{B6D7A8}1.00 & \cellcolor[HTML]{CFE5C5}0.67 & \cellcolor[HTML]{FFFFFF}0.00    & \cellcolor[HTML]{CFE5C5}0.67 \\
                       & human-avg                        & \cellcolor[HTML]{FFFFFF}0.00    & \cellcolor[HTML]{FFFFFF}0.00    & \cellcolor[HTML]{B6D7A8}1.00 & \cellcolor[HTML]{CFE5C5}0.67 & \cellcolor[HTML]{B6D7A8}1.00 & \cellcolor[HTML]{E7F2E2}0.33 & \cellcolor[HTML]{FFFFFF}0.00    & \cellcolor[HTML]{B6D7A8}1.00 & \cellcolor[HTML]{FFFFFF}0.00    & \cellcolor[HTML]{B6D7A8}1.00 & \cellcolor[HTML]{FFFFFF}0.00    & \cellcolor[HTML]{B6D7A8}1.00 \\
                       & \cellcolor[HTML]{FFFFFF}human-comet & \cellcolor[HTML]{FFFFFF}0.00    & \cellcolor[HTML]{E7F2E2}0.33 & \cellcolor[HTML]{FFFFFF}0.00    & \cellcolor[HTML]{B6D7A8}1.00 & \cellcolor[HTML]{FFFFFF}0.00    & \cellcolor[HTML]{CFE5C5}0.67 & \cellcolor[HTML]{FFFFFF}0.00    & \cellcolor[HTML]{B6D7A8}1.00 & \cellcolor[HTML]{FFFFFF}0.00    & \cellcolor[HTML]{B6D7A8}1.00 & \cellcolor[HTML]{FFFFFF}0.00    & \cellcolor[HTML]{B6D7A8}1.00 \\
 \midrule

\multirow{3}{*}{\STAB{\rotatebox[origin=c]{90}{\small \acs{EXP3}}}} & human-zero                            & \cellcolor[HTML]{FFFFFF}0.00    & \cellcolor[HTML]{E7F2E2}0.33 & \cellcolor[HTML]{FFFFFF}0.00    & \cellcolor[HTML]{FFFFFF}0.00    & \cellcolor[HTML]{FFFFFF}0.00    & \cellcolor[HTML]{FFFFFF}0.00    & \cellcolor[HTML]{FFFFFF}0.00    & \cellcolor[HTML]{E7F2E2}0.33 & \cellcolor[HTML]{B6D7A8}1.00 & \cellcolor[HTML]{E7F2E2}0.33 & \cellcolor[HTML]{B6D7A8}1.00 & \cellcolor[HTML]{E7F2E2}0.33 \\
                       & human-avg                        & \cellcolor[HTML]{FFFFFF}0.00    & \cellcolor[HTML]{FFFFFF}0.00    & \cellcolor[HTML]{FFFFFF}0.00    & \cellcolor[HTML]{E7F2E2}0.33 & \cellcolor[HTML]{FFFFFF}0.00    & \cellcolor[HTML]{CFE5C5}0.67 & \cellcolor[HTML]{FFFFFF}0.00    & \cellcolor[HTML]{E7F2E2}0.33 & \cellcolor[HTML]{FFFFFF}0.00    & \cellcolor[HTML]{E7F2E2}0.33 & \cellcolor[HTML]{FFFFFF}0.00    & \cellcolor[HTML]{E7F2E2}0.33 \\
                       & \cellcolor[HTML]{FFFFFF}human-comet & \cellcolor[HTML]{FFFFFF}0.00    & \cellcolor[HTML]{FFFFFF}0.00    & \cellcolor[HTML]{FFFFFF}0.00    & \cellcolor[HTML]{FFFFFF}0.00    & \cellcolor[HTML]{FFFFFF}0.00    & \cellcolor[HTML]{E7F2E2}0.33 & \cellcolor[HTML]{FFFFFF}0.00    & \cellcolor[HTML]{FFFFFF}0.00    & \cellcolor[HTML]{FFFFFF}0.00    & \cellcolor[HTML]{FFFFFF}0.00    & \cellcolor[HTML]{FFFFFF}0.00    & \cellcolor[HTML]{E7F2E2}0.33 \\
 \bottomrule
\end{tabular}
\caption{Overlap ratios of top 1 and top 3 systems in common between the online approaches and the official ranking for {\tt en-de}. Recall that, for this pair, the official ranking differed depending on whether the z-score or the raw score was considered.}
\label{tab:ende}
\end{table*}

\begin{table*}[t]
\centering
\begin{tabular}{@{}clrrrrrrrrrrrr@{}}
\toprule
\multicolumn{2}{r}{\textbf{Iteration}}                    & \multicolumn{2}{c}{10}                                      & \multicolumn{2}{c}{50}                                      & \multicolumn{2}{c}{100}                                     & \multicolumn{2}{c}{500}                                     & \multicolumn{2}{c}{1000}                                    & \multicolumn{2}{c}{1701}                                    \\ \midrule
\multicolumn{2}{r}{\textbf{Top}}                          & \multicolumn{1}{c}{1}        & \multicolumn{1}{c}{3}        & \multicolumn{1}{c}{1}        & \multicolumn{1}{c}{3}        & \multicolumn{1}{c}{1}        & \multicolumn{1}{c}{3}        & \multicolumn{1}{c}{1}        & \multicolumn{1}{c}{3}        & \multicolumn{1}{c}{1}        & \multicolumn{1}{c}{3}        & \multicolumn{1}{c}{1}        & \multicolumn{1}{c}{3}        \\ \midrule

\multirow{3}{*}{\STAB{\rotatebox[origin=c]{90}{\small \acs{EWAF}}}} & human-zero                            & \cellcolor[HTML]{FFFFFF}0.00    & \cellcolor[HTML]{CFE5C5}0.67 & \cellcolor[HTML]{FFFFFF}0.00    & \cellcolor[HTML]{E7F2E2}0.33 & \cellcolor[HTML]{FFFFFF}0.00    & \cellcolor[HTML]{E7F2E2}0.33 & \cellcolor[HTML]{FFFFFF}0.00    & \cellcolor[HTML]{CFE5C5}0.67 & \cellcolor[HTML]{FFFFFF}0.00    & \cellcolor[HTML]{CFE5C5}0.67 & \cellcolor[HTML]{B6D7A8}1.00 & \cellcolor[HTML]{CFE5C5}0.67 \\
                       & human-avg                        & \cellcolor[HTML]{FFFFFF}0.00    & \cellcolor[HTML]{CFE5C5}0.67 & \cellcolor[HTML]{FFFFFF}0.00    & \cellcolor[HTML]{CFE5C5}0.67 & \cellcolor[HTML]{FFFFFF}0.00    & \cellcolor[HTML]{CFE5C5}0.67 & \cellcolor[HTML]{FFFFFF}0.00    & \cellcolor[HTML]{E7F2E2}0.33 & \cellcolor[HTML]{FFFFFF}0.00    & \cellcolor[HTML]{E7F2E2}0.33 & \cellcolor[HTML]{FFFFFF}0.00    & \cellcolor[HTML]{E7F2E2}0.33 \\
                       & \cellcolor[HTML]{FFFFFF}human-comet & \cellcolor[HTML]{B6D7A8}1.00 & \cellcolor[HTML]{B6D7A8}1.00 & \cellcolor[HTML]{B6D7A8}1.00 & \cellcolor[HTML]{CFE5C5}0.67 & \cellcolor[HTML]{B6D7A8}1.00 & \cellcolor[HTML]{B6D7A8}1.00 & \cellcolor[HTML]{B6D7A8}1.00 & \cellcolor[HTML]{B6D7A8}1.00 & \cellcolor[HTML]{B6D7A8}1.00 & \cellcolor[HTML]{B6D7A8}1.00 & \cellcolor[HTML]{B6D7A8}1.00 & \cellcolor[HTML]{B6D7A8}1.00 \\
 \midrule

\multirow{3}{*}{\STAB{\rotatebox[origin=c]{90}{\small \acs{EXP3}}}} & human-zero                            & \cellcolor[HTML]{FFFFFF}0.00    & \cellcolor[HTML]{CFE5C5}0.67 & \cellcolor[HTML]{FFFFFF}0.00    & \cellcolor[HTML]{E7F2E2}0.33 & \cellcolor[HTML]{FFFFFF}0.00    & \cellcolor[HTML]{E7F2E2}0.33 & \cellcolor[HTML]{FFFFFF}0.00    & \cellcolor[HTML]{E7F2E2}0.33 & \cellcolor[HTML]{B6D7A8}1.00 & \cellcolor[HTML]{CFE5C5}0.67 & \cellcolor[HTML]{B6D7A8}1.00 & \cellcolor[HTML]{CFE5C5}0.67 \\
                       & human-avg                        & \cellcolor[HTML]{FFFFFF}0.00    & \cellcolor[HTML]{E7F2E2}0.33 & \cellcolor[HTML]{FFFFFF}0.00    & \cellcolor[HTML]{E7F2E2}0.33 & \cellcolor[HTML]{FFFFFF}0.00    & \cellcolor[HTML]{E7F2E2}0.33 & \cellcolor[HTML]{FFFFFF}0.00    & \cellcolor[HTML]{E7F2E2}0.33 & \cellcolor[HTML]{FFFFFF}0.00    & \cellcolor[HTML]{E7F2E2}0.33 & \cellcolor[HTML]{FFFFFF}0.00    & \cellcolor[HTML]{E7F2E2}0.33 \\
                       & \cellcolor[HTML]{FFFFFF}human-comet & \cellcolor[HTML]{FFFFFF}0.00    & \cellcolor[HTML]{E7F2E2}0.33 & \cellcolor[HTML]{FFFFFF}0.00    & \cellcolor[HTML]{CFE5C5}0.67 & \cellcolor[HTML]{FFFFFF}0.00    & \cellcolor[HTML]{CFE5C5}0.67 & \cellcolor[HTML]{FFFFFF}0.00    & \cellcolor[HTML]{E7F2E2}0.33 & \cellcolor[HTML]{FFFFFF}0.00    & \cellcolor[HTML]{CFE5C5}0.67 & \cellcolor[HTML]{FFFFFF}0.00    & \cellcolor[HTML]{CFE5C5}0.67 \\
\bottomrule

\end{tabular}
\caption{Overlap ratios of top 1 and top 3 systems in common between the online approaches and the official ranking for {\tt fr-de}. Recall that this was the language pair with the lowest coverage of human assessments.}
\label{tab:frde}
\end{table*}

\begin{table*}[!ht]
\centering
\begin{tabular}{@{}clrrrrrrrrrrrr@{}}
\toprule
\multicolumn{2}{r}{\textbf{Iteration}}                    & \multicolumn{2}{c}{10}                                      & \multicolumn{2}{c}{50}                                      & \multicolumn{2}{c}{100}                                     & \multicolumn{2}{c}{500}                                     & \multicolumn{2}{c}{1000}                                    & \multicolumn{2}{c}{1997}                                    \\ \midrule
\multicolumn{2}{r}{\textbf{Top}}                          & \multicolumn{1}{c}{1}        & \multicolumn{1}{c}{3}        & \multicolumn{1}{c}{1}        & \multicolumn{1}{c}{3}        & \multicolumn{1}{c}{1}        & \multicolumn{1}{c}{3}        & \multicolumn{1}{c}{1}        & \multicolumn{1}{c}{3}        & \multicolumn{1}{c}{1}        & \multicolumn{1}{c}{3}        & \multicolumn{1}{c}{1}        & \multicolumn{1}{c}{3}        \\ \midrule

\multirow{3}{*}{\STAB{\rotatebox[origin=c]{90}{\small \acs{EWAF}}}}  & human-zero                            & \cellcolor[HTML]{FFFFFF}0.00    & \cellcolor[HTML]{E7F2E2}0.33 & \cellcolor[HTML]{B6D7A8}1.00 & \cellcolor[HTML]{CFE5C5}0.67 & \cellcolor[HTML]{B6D7A8}1.00 & \cellcolor[HTML]{CFE5C5}0.67 & \cellcolor[HTML]{FFFFFF}0.00    & \cellcolor[HTML]{CFE5C5}0.67 & \cellcolor[HTML]{B6D7A8}1.00 & \cellcolor[HTML]{B6D7A8}1.00 & \cellcolor[HTML]{FFFFFF}0.00    & \cellcolor[HTML]{B6D7A8}1.00 \\
                       & human-avg                        & \cellcolor[HTML]{FFFFFF}0.00    & \cellcolor[HTML]{E7F2E2}0.33 & \cellcolor[HTML]{FFFFFF}0.00    & \cellcolor[HTML]{E7F2E2}0.33 & \cellcolor[HTML]{B6D7A8}1.00 & \cellcolor[HTML]{CFE5C5}0.67 & \cellcolor[HTML]{B6D7A8}1.00 & \cellcolor[HTML]{CFE5C5}0.67 & \cellcolor[HTML]{B6D7A8}1.00 & \cellcolor[HTML]{CFE5C5}0.67 & \cellcolor[HTML]{B6D7A8}1.00 & \cellcolor[HTML]{B6D7A8}1.00 \\
                       & \cellcolor[HTML]{FFFFFF}human-comet & \cellcolor[HTML]{FFFFFF}0.00    & \cellcolor[HTML]{E7F2E2}0.33 & \cellcolor[HTML]{B6D7A8}1.00 & \cellcolor[HTML]{CFE5C5}0.67 & \cellcolor[HTML]{B6D7A8}1.00 & \cellcolor[HTML]{CFE5C5}0.67 & \cellcolor[HTML]{B6D7A8}1.00 & \cellcolor[HTML]{B6D7A8}1.00 & \cellcolor[HTML]{B6D7A8}1.00 & \cellcolor[HTML]{B6D7A8}1.00 & \cellcolor[HTML]{B6D7A8}1.00 & \cellcolor[HTML]{B6D7A8}1.00 \\
\midrule

\multirow{3}{*}{\STAB{\rotatebox[origin=c]{90}{\small \acs{EXP3}}}} & human-zero                            & \cellcolor[HTML]{FFFFFF}0.00    & \cellcolor[HTML]{E7F2E2}0.33 & \cellcolor[HTML]{FFFFFF}0.00    & \cellcolor[HTML]{CFE5C5}0.67 & \cellcolor[HTML]{FFFFFF}0.00    & \cellcolor[HTML]{CFE5C5}0.67 & \cellcolor[HTML]{FFFFFF}0.00    & \cellcolor[HTML]{CFE5C5}0.67 & \cellcolor[HTML]{FFFFFF}0.00    & \cellcolor[HTML]{CFE5C5}0.67 & \cellcolor[HTML]{B6D7A8}1.00 & \cellcolor[HTML]{B6D7A8}1.00 \\
                       & human-avg                        & \cellcolor[HTML]{FFFFFF}0.00    & \cellcolor[HTML]{E7F2E2}0.33 & \cellcolor[HTML]{FFFFFF}0.00    & \cellcolor[HTML]{E7F2E2}0.33 & \cellcolor[HTML]{FFFFFF}0.00    & \cellcolor[HTML]{E7F2E2}0.33 & \cellcolor[HTML]{FFFFFF}0.00    & \cellcolor[HTML]{CFE5C5}0.67 & \cellcolor[HTML]{B6D7A8}1.00 & \cellcolor[HTML]{CFE5C5}0.67 & \cellcolor[HTML]{B6D7A8}1.00 & \cellcolor[HTML]{CFE5C5}0.67 \\
                       & \cellcolor[HTML]{FFFFFF}human-comet & \cellcolor[HTML]{FFFFFF}0.00    & \cellcolor[HTML]{E7F2E2}0.33 & \cellcolor[HTML]{B6D7A8}1.00 & \cellcolor[HTML]{E7F2E2}0.33 & \cellcolor[HTML]{B6D7A8}1.00 & \cellcolor[HTML]{E7F2E2}0.33 & \cellcolor[HTML]{B6D7A8}1.00 & \cellcolor[HTML]{CFE5C5}0.67 & \cellcolor[HTML]{B6D7A8}1.00 & \cellcolor[HTML]{CFE5C5}0.67 & \cellcolor[HTML]{B6D7A8}1.00 & \cellcolor[HTML]{CFE5C5}0.67 \\
\bottomrule
\end{tabular}
\caption{Overlap ratios of top 1 and top 3 systems in common between the online approaches and the official ranking for {\tt de-cs}.}
\label{tab:decs}
\end{table*}

\begin{table*}[t]
\centering
\begin{tabular}{@{}clrrrrrrrrrrrr@{}}
\toprule
\multicolumn{2}{r}{\textbf{Iteration}}                    & \multicolumn{2}{c}{10}                                      & \multicolumn{2}{c}{50}                                      & \multicolumn{2}{c}{100}                                     & \multicolumn{2}{c}{500}                                     & \multicolumn{2}{c}{1000}                                    & \multicolumn{2}{c}{1016}                                    \\ \midrule
\multicolumn{2}{r}{\textbf{Top}}                          & \multicolumn{1}{c}{1}        & \multicolumn{1}{c}{3}        & \multicolumn{1}{c}{1}        & \multicolumn{1}{c}{3}        & \multicolumn{1}{c}{1}        & \multicolumn{1}{c}{3}        & \multicolumn{1}{c}{1}        & \multicolumn{1}{c}{3}        & \multicolumn{1}{c}{1}        & \multicolumn{1}{c}{3}        & \multicolumn{1}{c}{1}        & \multicolumn{1}{c}{3}        \\ \midrule

\multirow{2}{*}{\STAB{\rotatebox[origin=c]{90}{\small \acs{EWAF}}}}   & human-zero                            & \cellcolor[HTML]{B6D7A8}1.00 & \cellcolor[HTML]{CFE5C5}0.67 & \cellcolor[HTML]{B6D7A8}1.00 & \cellcolor[HTML]{B6D7A8}1.00 & \cellcolor[HTML]{B6D7A8}1.00 & \cellcolor[HTML]{B6D7A8}1.00 & \cellcolor[HTML]{B6D7A8}1.00 & \cellcolor[HTML]{CFE5C5}0.67 & \cellcolor[HTML]{B6D7A8}1.00 & \cellcolor[HTML]{CFE5C5}0.67 & \cellcolor[HTML]{B6D7A8}1.00 & \cellcolor[HTML]{CFE5C5}0.67 \\
                       & \cellcolor[HTML]{FFFFFF}human-comet & \cellcolor[HTML]{B6D7A8}1.00 & \cellcolor[HTML]{CFE5C5}0.67 & \cellcolor[HTML]{FFFFFF}0.00    & \cellcolor[HTML]{CFE5C5}0.67 & \cellcolor[HTML]{FFFFFF}0.00    & \cellcolor[HTML]{CFE5C5}0.67 & \cellcolor[HTML]{FFFFFF}0.00    & \cellcolor[HTML]{CFE5C5}0.67 & \cellcolor[HTML]{FFFFFF}0.00    & \cellcolor[HTML]{E7F2E2}0.33 & \cellcolor[HTML]{FFFFFF}0.00    & \cellcolor[HTML]{E7F2E2}0.33 \\
\midrule

\multirow{2}{*}{\STAB{\rotatebox[origin=c]{90}{\small \acs{EXP3}}}}   & human-zero                            & \cellcolor[HTML]{FFFFFF}0.00    & \cellcolor[HTML]{E7F2E2}0.33 & \cellcolor[HTML]{FFFFFF}0.00    & \cellcolor[HTML]{CFE5C5}0.67 & \cellcolor[HTML]{FFFFFF}0.00    & \cellcolor[HTML]{E7F2E2}0.33 & \cellcolor[HTML]{FFFFFF}0.00    & \cellcolor[HTML]{E7F2E2}0.33 & \cellcolor[HTML]{FFFFFF}0.00    & \cellcolor[HTML]{E7F2E2}0.33 & \cellcolor[HTML]{FFFFFF}0.00    & \cellcolor[HTML]{E7F2E2}0.33 \\
                       & \cellcolor[HTML]{FFFFFF}human-comet & \cellcolor[HTML]{FFFFFF}0.00    & \cellcolor[HTML]{E7F2E2}0.33 & \cellcolor[HTML]{B6D7A8}1.00 & \cellcolor[HTML]{E7F2E2}0.33 & \cellcolor[HTML]{B6D7A8}1.00 & \cellcolor[HTML]{E7F2E2}0.33 & \cellcolor[HTML]{B6D7A8}1.00 & \cellcolor[HTML]{E7F2E2}0.33 & \cellcolor[HTML]{FFFFFF}0.00    & \cellcolor[HTML]{CFE5C5}0.67 & \cellcolor[HTML]{FFFFFF}0.00    & \cellcolor[HTML]{CFE5C5}0.67 \\

\bottomrule
\end{tabular}
\caption{Overlap ratios of top 1 and top 3 systems in common between the online approaches and the official ranking for {\tt gu-en}. Recall that there were three systems competing on this language pair that did not receive human assessments at all (thus, using {\tt human-avg} yields the same results as using {\tt human-zero}).}
\label{tab:guen}
\end{table*}

\begin{table*}[!ht]
\centering
\begin{tabular}{@{}clrrrrrrrrrr@{}}
\toprule
\multicolumn{2}{r}{\textbf{Iteration}} & \multicolumn{2}{c}{10}                                      & \multicolumn{2}{c}{50}                                   & \multicolumn{2}{c}{100}                                  & \multicolumn{2}{c}{500}                                     & \multicolumn{2}{c}{1000}                                    \\ \midrule
\multicolumn{2}{r}{\textbf{Top}}       & \multicolumn{1}{c}{1}        & \multicolumn{1}{c}{3}        & \multicolumn{1}{c}{1}     & \multicolumn{1}{c}{3}        & \multicolumn{1}{c}{1}     & \multicolumn{1}{c}{3}        & \multicolumn{1}{c}{1}        & \multicolumn{1}{c}{3}        & \multicolumn{1}{c}{1}        & \multicolumn{1}{c}{3}        \\ \midrule

                    {\small \acs{EWAF}}       & human-zero     & \cellcolor[HTML]{FFFFFF}0.00    & \cellcolor[HTML]{FFFFFF}0.00    & \cellcolor[HTML]{FFFFFF}0.00 & \cellcolor[HTML]{CFE5C5}0.67 & \cellcolor[HTML]{FFFFFF}0.00 & \cellcolor[HTML]{CFE5C5}0.67 & \cellcolor[HTML]{B6D7A8}1.00 & \cellcolor[HTML]{CFE5C5}0.67 & \cellcolor[HTML]{FFFFFF}0.00    & \cellcolor[HTML]{B6D7A8}1.00 \\
\midrule

                     {\small \acs{EXP3}}      & human-zero     & \cellcolor[HTML]{FFFFFF}0.00    & \cellcolor[HTML]{E7F2E2}0.33 & \cellcolor[HTML]{FFFFFF}0.00 & \cellcolor[HTML]{FFFFFF}0.00    & \cellcolor[HTML]{FFFFFF}0.00 & \cellcolor[HTML]{E7F2E2}0.33 & \cellcolor[HTML]{B6D7A8}1.00 & \cellcolor[HTML]{CFE5C5}0.67 & \cellcolor[HTML]{B6D7A8}1.00 & \cellcolor[HTML]{CFE5C5}0.67 \\
\bottomrule

\end{tabular}
\caption{Overlap ratios of top 1 and top 3 systems in common between the online approaches and the official ranking for {\tt lt-en}. Recall that this was the only language pair for which all the translations received at least one human assessment, thus there is no need to use a fallback loss function.}
\label{tab:lten}
\end{table*}

In order to observe whether (and how soon) our online approach converges to the best systems, we report the overlap between the top $n = 1, 3$ systems with greatest weights according to our approaches, $\hat{s_n}$, and the top $n = 1, 3$ systems according to the shared task's official ranking, $s_n^*$, at specific iterations:

\begin{equation}
top_n = \frac{| \hat{s_n} \cap s_n^*|}{n} , n = 1, 3
\label{eq:tops}
\end{equation}

We preferred this metric over a rank correlation metric, as we are focused on whether our online approach follows the performance of the best \ac{MT} systems. In a realistic scenario (e.g., a Web MT service), a user would most likely rely solely on the main translation returned, or would at most consider one or two alternative translations. 
Moreover, due to the lack of a large enough coverage of human assessments, the scores obtained in the shared task are not reliable enough to discriminate between similarly performing systems.

Starting with {\tt en-de} (Table~\ref{tab:ende}), this was the language pair for which our approach appears to be the least successful, since, for most of the iterations examined, it failed to converge to the best system. Even so, it managed to converge to the top 3 systems, doing so particularly early in the learning process (50 iterations) when using \ac{EWAF} with {\tt human-avg} and {\tt human-comet} as loss functions (i.e., when using fallback scores). Recall that, for this language pair, there were different official winning systems depending on whether one considers the z-score or the raw score (recall Table~\ref{tab:top3systems}); since we use the raw score as the loss function, it is expectable that our approach does not necessarily converge to the winner according to the z-score. 

For {\tt fr-de} (Table~\ref{tab:frde}), our online approach often converges to the top 3 systems (or a subset of them) throughout the learning process (even at just 10 iterations), and it also converges to the best system when using \ac{EWAF} with {\tt human-comet}. This is a particularly interesting result if we recall that, out of the five pairs considered, {\tt fr-de} had the lowest coverage of human assessments by far (see Table~\ref{tab:testsets}), thus suggesting that using {\sc Comet} may be an adequate fallback strategy. 

\begin{figure*}[t]
\centering
\pdftooltip{
\includegraphics[width=1\linewidth]{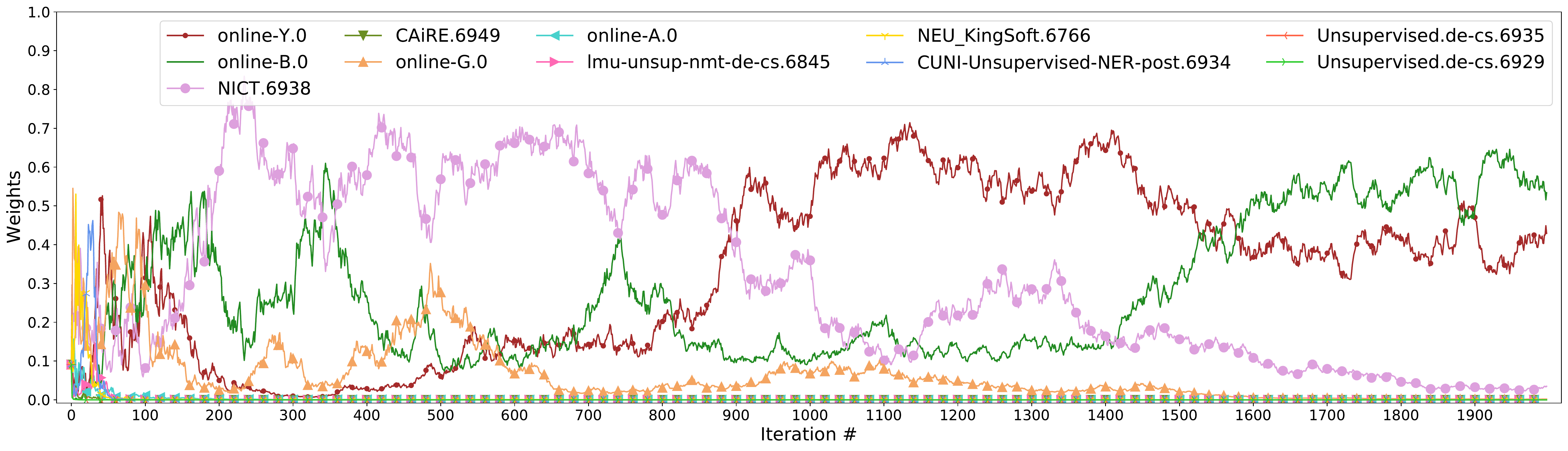}}
{Weight evolution curves for each machine translation system, over the course of nineteen ninety seven iterations. There are eleven systems: online Y, online B, NICT, CAIRE, online G, online A, lmu unsupervised nmt, NEU, CUNI, Unsupervised 6935, and Unsupervised 6929. Out of these, online Y, online B, and NICT have the greater weight values, alternating among them.}
\caption{Weight evolution per \ac{MT} system when using \ac{EWAF} and {\tt human-zero} as the loss function ({\tt de-cs}). Recall that, for this language pair, the official top 3 systems were online-Y, online-B, and NICT.}
\label{fig:weightsEWAFhuman}
\end{figure*}

\begin{figure*}[t]
\centering
\pdftooltip{
\includegraphics[width=1\linewidth]{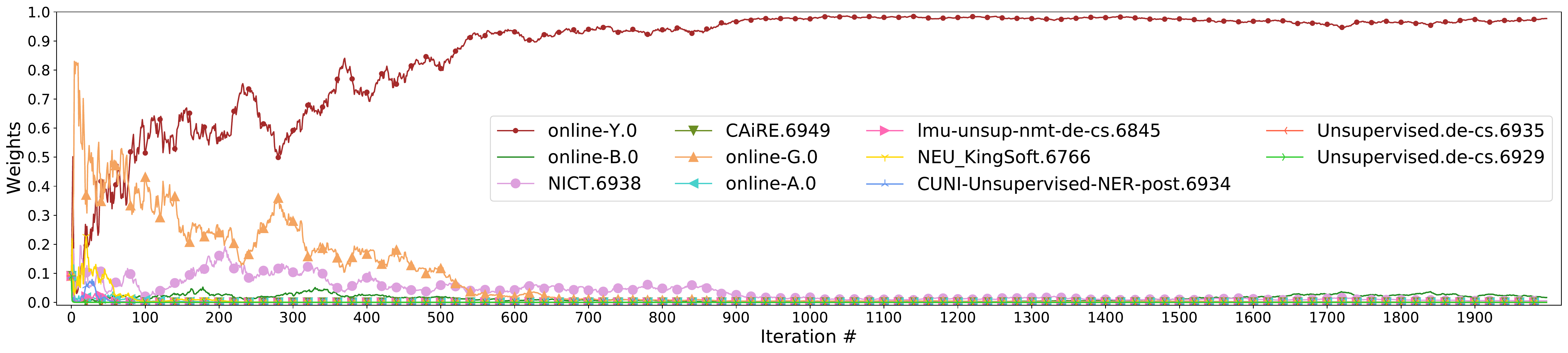}}
{Weight evolution curves for each machine translation system, over the course of nineteen ninety seven iterations. There are eleven systems: online Y, online B, NICT, CAIRE, online G, online A, lmu unsupervised nmt, NEU, CUNI, Unsupervised 6935, and Unsupervised 6929. Out of these, online Y has the greater weight values for most of the iterations, starting before the iteration number one hundred.}
\caption{Weight evolution per \ac{MT} system when using \ac{EWAF} and {\tt human-avg} as the loss function ({\tt de-cs}).}
\label{fig:weightsEWAFhumanavg}
\end{figure*}

\begin{figure*}[!ht]
\centering
\pdftooltip{
\includegraphics[width=1\linewidth]{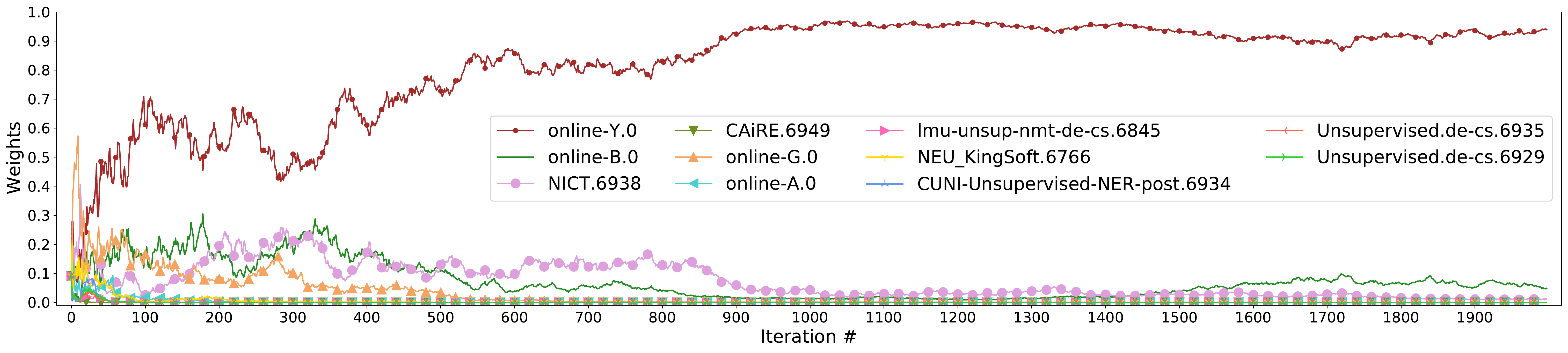}}
{Weight evolution curves for each machine translation system, over the course of nineteen ninety seven iterations. There are eleven systems: online Y, online B, NICT, CAIRE, online G, online A, lmu unsupervised nmt, NEU, CUNI, Unsupervised 6935, and Unsupervised 6929. Out of these, online Y has the greater weight values for most of the iterations, starting before the iteration number fifty.}
\caption{Weight evolution per \ac{MT} system when using \ac{EWAF} and {\tt human-comet} as the loss function ({\tt de-cs}).}
\label{fig:weightsEWAFhumancomet}
\end{figure*}

\begin{figure*}[!ht]
\centering
\pdftooltip{
\includegraphics[width=1\linewidth]{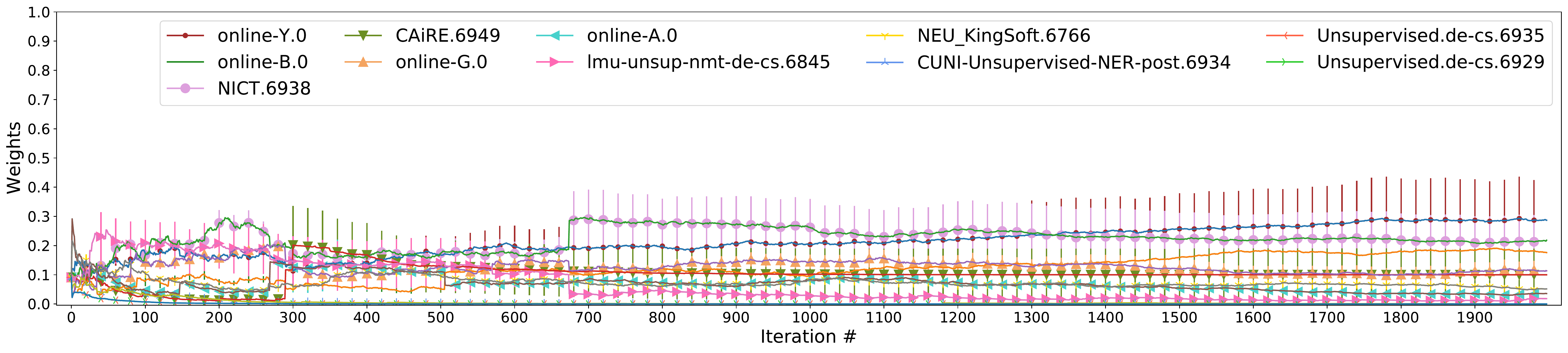}}
{Weight evolution curves for each machine translation system, over the course of nineteen ninety seven iterations. There are eleven systems: online Y, online B, NICT, CAIRE, online G, online A, lmu unsupervised nmt, NEU, CUNI, Unsupervised 6935, and Unsupervised 6929. Out of these, online Y, NICT, and online B have the greater weight values, but the weight distribution is very balanced across systems, and there is a large variance across runs.}
\caption{Weight evolution per \ac{MT} system when using \ac{EXP3} and {\tt human-zero} as the loss function ({\tt de-cs}), averaged across 10 runs (the error bars represent the weights' variance across the 10 runs).}
\label{fig:weightsEXP3human}
\end{figure*}

For {\tt de-cs} (Table~\ref{tab:decs}), we can see that, regardless of the algorithm and loss function used, there is an overlap of at least one system between our top 3 and the shared task's official top 3, after going through only as few as 10 iterations (despite a considerable lack of human assessments in this language pair). We can also see that the {\tt human-comet} loss function is the most successful overall, which reinforces the idea that {\sc Comet} may be an appropriate fallback metric in the absence of human scores for a given translation. Since this is the language pair for which there seems to be a more similar performance across different algorithms and loss functions, we also report the weight evolution plots for this pair in order to inspect what changes depending on the algorithm and fallback strategy used\footnote{The plots for the remaining pairs can be found in App.~\ref{sec:appendixA}.}. Looking at \ac{EWAF} combined with the {\tt human-zero} loss function (Fig.~\ref{fig:weightsEWAFhuman}), one can see a rather irregular evolution for the weights of the top systems, which may be explained by the distribution of the translations lacking human assessments across different systems and learning iterations. Using the {\tt human-avg} loss function (Fig.\ref{fig:weightsEWAFhumanavg}) allows for a more monotonous evolution, by rewarding the systems that were doing better overall in the absence of human assessments. Using the {\tt human-comet} loss function (Fig.~\ref{fig:weightsEWAFhumancomet}) paints a similar picture, as the {\sc Comet} scores for this language pair seem to be in line with the official ranking (although they appear to benefit the third best system in detriment of the second best). Finally, using \ac{EXP3} instead of \ac{EWAF} (Fig.~\ref{fig:weightsEXP3human}), combined with {\tt human-zero}, leads to much less pronounced weights, but still in line with the official ranking. Recall that, for \ac{EXP3}, these weights are averaged across different runs: since each run may lead to different top systems, the difference between the averaged weights ends up being more smooth, i.e., there is a great variance across runs (this happens regardless of the language pair or loss function).

As for {\tt gu-en} (Table~\ref{tab:guen}), our approach (using \ac{EWAF} with {\tt human-zero}) converges to the best system and to a subset of the top 3 within just 10 iterations; on the other hand, using {\tt human-comet} does not do as well as not using a fallback strategy, at least when combined with \ac{EWAF}. However, recall that, for this pair, there were systems that did not receive any human assessments at all for their translations (that being the reason why we do not report {\tt human-avg} for this pair: the resulting weights end up being the same as when using {\tt human-zero}). One of the systems that did not receive any human assessments, {\tt online-B}, ended up receiving high {\sc Comet} scores, thus leading to a weaker overlap between the online approach ranking and the official ranking. 

Finally, for {\tt lt-en} (Table~\ref{tab:lten}) we only report the {\tt human-zero} loss function, since this is the only pair for which there are human assessments for all translations. Interestingly, the online approaches do not do well as quickly as for other pairs, but eventually get there (within 100 to 500 iterations). 

To sum up these results: although factors like the coverage of human assessments or the combinations of online algorithm and loss function used influence how well our approach does, we can still conclude that using an online learning approach allows to converge to the top 3 systems according to the official ranking (or at least to a subset of them) in just a few hundred iterations (and, in some cases, in just a few dozens of iterations) for all the language pairs considered. 

%%%%%%%%%%%%%%%%%%%%%%%%%%%%%%%%%%%%%%%%%%%%%%%%%%%%%%%%%%%%%%%%%%%%%%%%%%%%%%%%
% Related Work
\section{Related work}
\label{sec:rw}

\subsection{\ac{WMT}'19 News Translation Shared Task}

Every year, since 2006, the Conference on Machine Translation (WMT) is responsible for organizing several shared tasks where participants push the limits of \ac{MT} and \ac{MT} evaluation \citep{koehn-monz-2006-manual,barrault-etal-2020-findings}. In the News Translation shared task, participants submit the outputs of their systems that are then evaluated by a community of human evaluators using Direct Assessment  scores \citep{graham-etal-2013-continuous}. Thus, the winner is the system that achieves the highest average score. 
For \ac{WMT}'19 \citep{Barrault2019}, most of the competing systems followed a Transformer architecture \citep{Vaswani2017}, with the main differences among them being: (i) whether they considered document-level or only sentence-level information; (ii) whether they were trained only on the training data provided by the shared task, or on additional sources as well; (iii) whether they consisted of a single model or an ensemble. 

\subsection{Online learning for \acl{MT}}

There has been a number of online learning approaches applied to \ac{MT} in the past, mainly in Interactive \ac{MT} and/or post-editing \ac{MT} systems. 
However, most approaches aim at learning the parameters or feature weights of an \ac{MT} model \citep{Mathur2013,Denkowski2014,Ortiz-Martinez2016,Sokolov2016,Nguyen2017,Lam2018} or fine-tuning a pre-trained model for domain adaptation \citep{Turchi2017,Karimova2018,Peris2019}. Even in cases where the \ac{MT} model is composed of several sub-models (e.g., \citet{Ortiz-Martinez2016}), the goal is to online learn each sub-model's specific parameters (while our learning goal is the weights of each system in an ensemble). 
Another key difference between these approaches and ours is that most of them use human post-edited translations as a source of feedback. The exceptions to this are the systems competing for \ac{WMT}'17 shared task on online bandit learning for \ac{MT} \citep{Sokolov2017}, as well as \citet{Lam2018}, who use (simulated) quality judgments.

The most similar proposal to ours is that of \citet{Naradowsky2020}, who ensemble different \ac{MT} systems and dynamically select the best one for a given \ac{MT} task or domain using stochastic multi-armed bandits and contextual bandits. The bandit algorithms learn from feedback simulated using a sentence-level {\sc Bleu} score between the selected automatic translation and a reference translation.

Thus, to the best of our knowledge, we are the first to frame the \ac{MT} problem as a problem of prediction with expert advice and adversarial multi-armed bandits in order to combine different systems into an ensemble that converges to the performance of the best individual systems, simulating the human-in-the-loop by using actual human assessments (when available).

%%%%%%%%%%%%%%%%%%%%%%%%%%%%%%%%%%%%%%%%%%%%%%%%%%%%%%%%%%%%%%%%%%%%%%%%%%%%%%%%
% CONCLUSIONS
\section{Conclusions and future work}
\label{sec:cfw}

We proposed an online learning approach to address the issue of finding the best \ac{MT} systems among an ensemble, while making the most of existing human feedback. In our experiments on \ac{WMT}'19 News Translation datasets, our approach converged to the top-3 systems (or a subset of them) according to the official shared task's ranking in just a few hundred iterations for all the language pairs considered (and just a few dozens in some cases), despite the lack of human assessments for many translations.
This is a promising result, not only for the purpose of reducing the human evaluations required to find the best systems in a shared task, but also for any \ac{MT} application that has access to an ensemble of multiple independent systems and to a source of feedback from which it can learn iteratively (e.g., Web translation services). 

Yet, our approach is limited by the {\em quality} of the collected human judgments. For future work, we plan to combine online learning with a more reliable human metric, such as the \ac{MQM} \citep{mqm}, so that we can focus on the quality of the assessments instead of their quantity. 

\section*{Acknowledgments}

This work was supported by: Fundação para a Ciência e a Tecnologia (FCT) under references UIDB/50021/2020 (INESC-ID multi-annual funding) and UIDB/00214/2020 (CLUL), as well as under the HOTSPOT project with reference PTDC/CCI-COM/7203/2020; Air Force Office of Scientific Research under award number FA9550-19-1-0020; P2020 program, supervised by Agência Nacional de Inovação (ANI), under the project CMU-PT Ref. 045909 (MAIA). 
Vânia Mendonça was funded by an FCT grant with reference SFRH/BD/121443/2016. 

The authors would like to thank the reviewers for their valuable comments, and to Soraia M. Alarcão for kindly proof-reading this document.  

%%%%%%%%%%%%%%%%%%%%%%%%%%%%%%%%%%%%%%%%%%%%%%%%%%%%%%%%%%%%%%%%%%%%%%%%%%%%%%%%
% REFS
\bibliographystyle{acl_natbib}
\bibliography{acl2021}

\begin{thebibliography}{39}
\expandafter\ifx\csname natexlab\endcsname\relax\def\natexlab#1{#1}\fi

\bibitem[{Auer et~al.(1995)Auer, Cesa-Bianchi, Freund, and Schapire}]{Auer1995}
Peter Auer, Nicolo Cesa-Bianchi, Yoav Freund, and Robert~E. Schapire. 1995.
\newblock \href {https://doi.org/10.1109/sfcs.1995.492488} {{Gambling in a
  rigged casino: the adversarial multi-armed bandit problem}}.
\newblock In \emph{Annual Symposium on Foundations of Computer Science -
  Proceedings}, pages 322--331.

\bibitem[{Barrault et~al.(2020)Barrault, Biesialska, Bojar, Costa-juss{\`a},
  Federmann, Graham, Grundkiewicz, Haddow, Huck, Joanis, Kocmi, Koehn, Lo,
  Ljube{\v{s}}i{\'c}, Monz, Morishita, Nagata, Nakazawa, Pal, Post, and
  Zampieri}]{barrault-etal-2020-findings}
Lo{\"\i}c Barrault, Magdalena Biesialska, Ond{\v{r}}ej Bojar, Marta~R.
  Costa-juss{\`a}, Christian Federmann, Yvette Graham, Roman Grundkiewicz,
  Barry Haddow, Matthias Huck, Eric Joanis, Tom Kocmi, Philipp Koehn, Chi-kiu
  Lo, Nikola Ljube{\v{s}}i{\'c}, Christof Monz, Makoto Morishita, Masaaki
  Nagata, Toshiaki Nakazawa, Santanu Pal, Matt Post, and Marcos Zampieri. 2020.
\newblock \href {https://www.aclweb.org/anthology/2020.wmt-1.1} {Findings of
  the 2020 conference on machine translation ({WMT}20)}.
\newblock In \emph{Proceedings of the Fifth Conference on Machine Translation},
  pages 1--55, Online. Association for Computational Linguistics.

\bibitem[{Barrault et~al.(2019)Barrault, Bojar, Costa-juss{\`{a}}, Federmann,
  Fishel, Graham, Haddow, Huck, Koehn, Malmasi, Monz, M{\"{u}}ller, Pal, Post,
  and Zampieri}]{Barrault2019}
Lo{\"{i}}c Barrault, Ondřej Bojar, Marta~R. Costa-juss{\`{a}}, Christian
  Federmann, Mark Fishel, Yvette Graham, Barry Haddow, Matthias Huck, Philipp
  Koehn, Shervin Malmasi, Christof Monz, Mathias M{\"{u}}ller, Santanu Pal,
  Matt Post, and Marcos Zampieri. 2019.
\newblock \href {https://doi.org/10.18653/v1/W19-5301} {{Findings of the 2019
  Conference on Machine Translation (WMT19)}}.
\newblock In \emph{Proceedings of the Fourth Conference on Machine Translation
  (Volume 2: Shared Task Papers, Day 1)}, pages 1--61, Stroudsburg, PA, USA.
  Association for Computational Linguistics.

\bibitem[{Bawden et~al.(2019)Bawden, Bogoychev, Germann, Grundkiewicz, Kirefu,
  {Miceli Barone}, and Birch}]{Bawden2019}
Rachel Bawden, Nikolay Bogoychev, Ulrich Germann, Roman Grundkiewicz, Faheem
  Kirefu, Antonio~Valerio {Miceli Barone}, and Alexandra Birch. 2019.
\newblock \href {https://doi.org/10.18653/v1/W19-5304} {{The University of
  Edinburgh's Submissions to the WMT19 News Translation Task}}.
\newblock In \emph{Proceedings of the Fourth Conference on Machine Translation
  (Volume 2: Shared Task Papers, Day 1)}, pages 103--115, Stroudsburg, PA, USA.
  Association for Computational Linguistics.

\bibitem[{Bei et~al.(2019)Bei, Zong, Yuan, Liu, and Fan}]{Bei2019}
Chao Bei, Hao Zong, Conghu Yuan, Qingming Liu, and Baoyong Fan. 2019.
\newblock \href {https://doi.org/10.18653/v1/W19-5305} {{GTCOM Neural Machine
  Translation Systems for WMT19}}.
\newblock In \emph{Proceedings of the Fourth Conference on Machine Translation
  (Volume 2: Shared Task Papers, Day 1)}, pages 116--121, Stroudsburg, PA, USA.
  Association for Computational Linguistics.

\bibitem[{Bojar et~al.(2017)Bojar, Chatterjee, Federmann, Graham, Haddow,
  Huang, Huck, Koehn, Liu, Logacheva, Monz, Negri, Post, Rubino, Specia, and
  Turchi}]{Bojar2017}
Ondřej Bojar, Rajen Chatterjee, Christian Federmann, Yvette Graham, Barry
  Haddow, Shujian Huang, Matthias Huck, Philipp Koehn, Qun Liu, Varvara
  Logacheva, Christof Monz, Matteo Negri, Matt Post, Raphael Rubino, Lucia
  Specia, and Marco Turchi. 2017.
\newblock \href {https://doi.org/10.18653/v1/W17-4717} {{Findings of the 2017
  Conference on Machine Translation (WMT17)}}.
\newblock In \emph{Proceedings of the Second Conference on Machine
  Translation}, pages 169--214, Stroudsburg, PA, USA. Association for
  Computational Linguistics.

\bibitem[{Bojar et~al.(2018)Bojar, Federmann, Fishel, Graham, Haddow, Koehn,
  and Monz}]{Bojar2018}
Ondřej Bojar, Christian Federmann, Mark Fishel, Yvette Graham, Barry Haddow,
  Philipp Koehn, and Christof Monz. 2018.
\newblock \href {https://doi.org/10.18653/v1/W18-6401} {{Findings of the 2018
  Conference on Machine Translation (WMT18)}}.
\newblock In \emph{Proceedings of the Third Conference on Machine Translation:
  Shared Task Papers}, pages 272--303, Stroudsburg, PA, USA. Association for
  Computational Linguistics.

\bibitem[{Bougares et~al.(2019)Bougares, Wottawa, Baillot, Barrault, and
  Bardet}]{Bougares2019}
Fethi Bougares, Jane Wottawa, Anne Baillot, Lo{\"{i}}c Barrault, and Adrien
  Bardet. 2019.
\newblock \href {https://doi.org/10.18653/v1/W19-5307} {{LIUM's Contributions
  to the WMT2019 News Translation Task: Data and Systems for German-French
  Language Pairs}}.
\newblock In \emph{Proceedings of the Fourth Conference on Machine Translation
  (Volume 2: Shared Task Papers, Day 1)}, pages 129--133, Stroudsburg, PA, USA.
  Association for Computational Linguistics.

\bibitem[{{Cesa-Bianchi} and Lugosi(2006)}]{cesa-bianchi06}
N.\ {Cesa-Bianchi} and G.\ Lugosi. 2006.
\newblock \emph{Prediction, Learning and Games}.
\newblock Cambridge University Press.

\bibitem[{Dabre et~al.(2019)Dabre, Chen, Marie, Wang, Fujita, Utiyama, and
  Sumita}]{Dabre2019}
Raj Dabre, Kehai Chen, Benjamin Marie, Rui Wang, Atsushi Fujita, Masao Utiyama,
  and Eiichiro Sumita. 2019.
\newblock \href {https://doi.org/10.18653/v1/W19-5313} {{NICT's Supervised
  Neural Machine Translation Systems for the WMT19 News Translation Task}}.
\newblock In \emph{Proceedings of the Fourth Conference on Machine Translation
  (Volume 2: Shared Task Papers, Day 1)}, pages 168--174, Stroudsburg, PA, USA.
  Association for Computational Linguistics.

\bibitem[{Denkowski et~al.(2014)Denkowski, Dyer, and Lavie}]{Denkowski2014}
Michael Denkowski, Chris Dyer, and Alon Lavie. 2014.
\newblock \href {https://doi.org/10.3115/v1/e14-1042} {{Learning from
  post-editing: Online model adaptation for statistical machine translation}}.
\newblock In \emph{Proceedings of the 14th Conference of the European Chapter
  of the Association for Computational Linguistics 2014}, pages 395--404.

\bibitem[{Freitag et~al.(2021)Freitag, Foster, Grangier, Ratnakar, Tan, and
  Macherey}]{Freitag2021}
Markus Freitag, George Foster, David Grangier, Viresh Ratnakar, Qijun Tan, and
  Wolfgang Macherey. 2021.
\newblock \href {http://arxiv.org/abs/2104.14478} {{Experts, Errors, and
  Context: A Large-Scale Study of Human Evaluation for Machine Translation}}.

\bibitem[{Graham et~al.(2013)Graham, Baldwin, Moffat, and
  Zobel}]{graham-etal-2013-continuous}
Yvette Graham, Timothy Baldwin, Alistair Moffat, and Justin Zobel. 2013.
\newblock \href {https://www.aclweb.org/anthology/W13-2305} {Continuous
  measurement scales in human evaluation of machine translation}.
\newblock In \emph{Proceedings of the 7th Linguistic Annotation Workshop and
  Interoperability with Discourse}, pages 33--41, Sofia, Bulgaria. Association
  for Computational Linguistics.

\bibitem[{Junczys-Dowmunt(2019)}]{Junczys-Dowmunt2019}
Marcin Junczys-Dowmunt. 2019.
\newblock \href {https://doi.org/10.18653/v1/W19-5321} {{Microsoft Translator
  at WMT 2019: Towards Large-Scale Document-Level Neural Machine Translation}}.
\newblock In \emph{Proceedings of the Fourth Conference on Machine Translation
  (Volume 2: Shared Task Papers, Day 1)}, pages 225--233, Stroudsburg, PA, USA.
  Association for Computational Linguistics.

\bibitem[{Karimova et~al.(2018)Karimova, Simianer, and Riezler}]{Karimova2018}
Sariya Karimova, Patrick Simianer, and Stefan Riezler. 2018.
\newblock \href {https://doi.org/10.1007/s10590-018-9224-8} {{A user-study on
  online adaptation of neural machine translation to human post-edits}}.
\newblock \emph{Machine Translation}, 32(4):309--324.

\bibitem[{Koehn and Monz(2006)}]{koehn-monz-2006-manual}
Philipp Koehn and Christof Monz. 2006.
\newblock \href {https://www.aclweb.org/anthology/W06-3114} {Manual and
  automatic evaluation of machine translation between {E}uropean languages}.
\newblock In \emph{Proceedings on the Workshop on Statistical Machine
  Translation}, pages 102--121, New York City. Association for Computational
  Linguistics.

\bibitem[{Lai and Robbins(1985)}]{Lai1985}
T.~L. Lai and Herbert Robbins. 1985.
\newblock \href {https://doi.org/10.1016/0196-8858(85)90002-8} {{Asymptotically
  efficient adaptive allocation rules}}.
\newblock \emph{Advances in Applied Mathematics}, 6(1):4--22.

\bibitem[{Lam et~al.(2018)Lam, Kreutzer, and Riezler}]{Lam2018}
Tsz~Kin Lam, Julia Kreutzer, and Stefan Riezler. 2018.
\newblock \href {http://arxiv.org/abs/1805.01553} {{A reinforcement learning
  approach to interactive-predictive neural machine translation}}.

\bibitem[{Li et~al.(2019)Li, Li, Xu, Lin, Liu, Liu, Wang, Zhang, Xu, Wang,
  Feng, Chen, Liu, Li, Wang, Xiao, and Zhu}]{Li2019}
Bei Li, Yinqiao Li, Chen Xu, Ye~Lin, Jiqiang Liu, Hui Liu, Ziyang Wang, Yuhao
  Zhang, Nuo Xu, Zeyang Wang, Kai Feng, Hexuan Chen, Tengbo Liu, Yanyang Li,
  Qiang Wang, Tong Xiao, and Jingbo Zhu. 2019.
\newblock \href {https://doi.org/10.18653/v1/W19-5325} {{The NiuTrans Machine
  Translation Systems for WMT19}}.
\newblock In \emph{Proceedings of the Fourth Conference on Machine Translation
  (Volume 2: Shared Task Papers, Day 1)}, pages 257--266, Stroudsburg, PA, USA.
  Association for Computational Linguistics.

\bibitem[{Lommel et~al.(2014)Lommel, Burchardt, and Uszkoreit}]{mqm}
Arle Lommel, Aljoscha Burchardt, and Hans Uszkoreit. 2014.
\newblock \href {https://doi.org/10.5565/rev/tradumatica.77} {Multidimensional
  quality metrics ({MQM}): A framework for declaring and describing translation
  quality metrics}.
\newblock \emph{Tradumàtica: tecnologies de la traducció}, 0:455--463.

\bibitem[{Ma et~al.(2019)Ma, Wei, Bojar, and Graham}]{ma-etal-2019-results}
Qingsong Ma, Johnny Wei, Ond{\v{r}}ej Bojar, and Yvette Graham. 2019.
\newblock \href {https://doi.org/10.18653/v1/W19-5302} {Results of the {WMT}19
  metrics shared task: Segment-level and strong {MT} systems pose big
  challenges}.
\newblock In \emph{Proceedings of the Fourth Conference on Machine Translation
  (Volume 2: Shared Task Papers, Day 1)}, pages 62--90, Florence, Italy.
  Association for Computational Linguistics.

\bibitem[{Mathur et~al.(2020)Mathur, Baldwin, and
  Cohn}]{mathur-etal-2020-tangled}
Nitika Mathur, Timothy Baldwin, and Trevor Cohn. 2020.
\newblock \href {https://doi.org/10.18653/v1/2020.acl-main.448} {Tangled up in
  {BLEU}: Reevaluating the evaluation of automatic machine translation
  evaluation metrics}.
\newblock In \emph{Proceedings of the 58th Annual Meeting of the Association
  for Computational Linguistics}, pages 4984--4997, Online. Association for
  Computational Linguistics.

\bibitem[{Mathur et~al.(2013)Mathur, Cettolo, and Federico}]{Mathur2013}
Prashant Mathur, Mauro Cettolo, and Marcello Federico. 2013.
\newblock {Online Learning Approaches in Computer Assisted Translation}.
\newblock In \emph{Proceedings of the Eighth Workshop on Statistical Machine
  Translation}, pages 301--308.

\bibitem[{Naradowsky et~al.(2020)Naradowsky, Zhang, and Duh}]{Naradowsky2020}
Jason Naradowsky, Xuan Zhang, and Kevin Duh. 2020.
\newblock \href {http://arxiv.org/abs/2002.09646} {{Machine Translation System
  Selection from Bandit Feedback}}.
\newblock In \emph{Proceedings of the 14th Conference of the Association for
  Machine Translation in the Americas}, pages 50--63.

\bibitem[{Ng et~al.(2019)Ng, Yee, Baevski, Ott, Auli, and Edunov}]{Ng2019}
Nathan Ng, Kyra Yee, Alexei Baevski, Myle Ott, Michael Auli, and Sergey Edunov.
  2019.
\newblock \href {https://doi.org/10.18653/v1/W19-5333} {{Facebook FAIR's WMT19
  News Translation Task Submission}}.
\newblock In \emph{Proceedings of the Fourth Conference on Machine Translation
  (Volume 2: Shared Task Papers, Day 1)}, pages 314--319, Stroudsburg, PA, USA.
  Association for Computational Linguistics.

\bibitem[{Nguyen et~al.(2017)Nguyen, {Daum{\'{e}} III}, and
  Boyd-Graber}]{Nguyen2017}
Khanh Nguyen, Hal {Daum{\'{e}} III}, and Jordan Boyd-Graber. 2017.
\newblock \href {https://doi.org/10.18653/v1/D17-1153} {{Reinforcement Learning
  for Bandit Neural Machine Translation with Simulated Human Feedback}}.
\newblock In \emph{Proceedings of the 2017 Conference on Empirical Methods in
  Natural Language Processing}, pages 1464--1474, Stroudsburg, PA, USA.
  Association for Computational Linguistics.

\bibitem[{Oravecz et~al.(2019)Oravecz, Bontcheva, Lardilleux, Tihanyi, and
  Eisele}]{Oravecz2019}
Csaba Oravecz, Katina Bontcheva, Adrien Lardilleux, L{\'{a}}szl{\'{o}} Tihanyi,
  and Andreas Eisele. 2019.
\newblock \href {https://doi.org/10.18653/v1/W19-5334} {{eTranslation's
  Submissions to the WMT 2019 News Translation Task}}.
\newblock In \emph{Proceedings of the Fourth Conference on Machine Translation
  (Volume 2: Shared Task Papers, Day 1)}, pages 320--326, Stroudsburg, PA, USA.
  Association for Computational Linguistics.

\bibitem[{Ortiz-Mart{\'{i}}nez(2016)}]{Ortiz-Martinez2016}
Daniel Ortiz-Mart{\'{i}}nez. 2016.
\newblock \href {https://doi.org/10.1162/COLI_a_00244} {{Online Learning for
  Statistical Machine Translation}}.
\newblock \emph{Computational Linguistics}, 42(1):121--161.

\bibitem[{Papineni et~al.(2002)Papineni, Roukos, Ward, and Zhu}]{papineni-bleu}
Kishore Papineni, Salim Roukos, Todd Ward, and Wei-Jing Zhu. 2002.
\newblock \href {https://doi.org/10.3115/1073083.1073135} {{B}leu: a method for
  automatic evaluation of machine translation}.
\newblock In \emph{Proceedings of the 40th Annual Meeting of the Association
  for Computational Linguistics}, pages 311--318, Philadelphia, Pennsylvania,
  USA. Association for Computational Linguistics.

\bibitem[{Peris and Casacuberta(2019)}]{Peris2019}
{\'{A}}lvaro Peris and Francisco Casacuberta. 2019.
\newblock \href {https://doi.org/10.1016/j.csl.2019.04.001} {{Online learning
  for effort reduction in interactive neural machine translation}}.
\newblock \emph{Computer Speech and Language}, 58:98--126.

\bibitem[{Pinnis et~al.(2019)Pinnis, Kri{\v{s}}lauks, and Rikters}]{Pinnis2019}
Marcis Pinnis, Rihards Kri{\v{s}}lauks, and Matīss Rikters. 2019.
\newblock \href {https://doi.org/10.18653/v1/W19-5335} {{Tilde's Machine
  Translation Systems for WMT 2019}}.
\newblock In \emph{Proceedings of the Fourth Conference on Machine Translation
  (Volume 2: Shared Task Papers, Day 1)}, pages 327--334, Stroudsburg, PA, USA.
  Association for Computational Linguistics.

\bibitem[{Rei et~al.(2020{\natexlab{a}})Rei, Stewart, Farinha, and
  Lavie}]{rei-etal-2020-comet}
Ricardo Rei, Craig Stewart, Ana~C Farinha, and Alon Lavie. 2020{\natexlab{a}}.
\newblock \href {https://doi.org/10.18653/v1/2020.emnlp-main.213} {{COMET}: A
  neural framework for {MT} evaluation}.
\newblock In \emph{Proceedings of the 2020 Conference on Empirical Methods in
  Natural Language Processing (EMNLP)}, pages 2685--2702, Online. Association
  for Computational Linguistics.

\bibitem[{Rei et~al.(2020{\natexlab{b}})Rei, Stewart, Farinha, and
  Lavie}]{rei-etal-2020-unbabels}
Ricardo Rei, Craig Stewart, Ana~C Farinha, and Alon Lavie. 2020{\natexlab{b}}.
\newblock \href {https://www.aclweb.org/anthology/2020.wmt-1.101} {Unbabel{'}s
  participation in the {WMT}20 metrics shared task}.
\newblock In \emph{Proceedings of the Fifth Conference on Machine Translation},
  pages 911--920, Online. Association for Computational Linguistics.

\bibitem[{Robbins(1952)}]{Herbert1952}
Herbert Robbins. 1952.
\newblock \href {https://doi.org/10.1090/S0002-9904-1952-09620-8} {{Some
  Aspects of the Sequential Design of Experiments}}.
\newblock \emph{Bulletin of the American Mathematical Society}, 58(5):527--535.

\bibitem[{Sokolov et~al.(2016)Sokolov, Kreutzer, Lo, and Riezler}]{Sokolov2016}
Artem Sokolov, Julia Kreutzer, Christopher Lo, and Stefan Riezler. 2016.
\newblock \href {https://doi.org/10.18653/v1/p16-1152} {{Learning structured
  predictors from bandit feedback for interactive NLP}}.
\newblock In \emph{Proceedings of the 54th Annual Meeting of the Association
  for Computational Linguistics}, pages 1610--1620.

\bibitem[{Sokolov et~al.(2017)Sokolov, Kreutzer, Sunderland, Danchenko,
  Szymaniak, F{\"{u}}rstenau, and Riezler}]{Sokolov2017}
Artem Sokolov, Julia Kreutzer, Kellen Sunderland, Pavel Danchenko, Witold
  Szymaniak, Hagen F{\"{u}}rstenau, and Stefan Riezler. 2017.
\newblock \href {https://doi.org/10.18653/v1/w17-4756} {{A Shared Task on
  Bandit Learning for Machine Translation}}.
\newblock In \emph{Proceedings of the Conference on Machine Translation (WMT)},
  volume~2, pages 514--524.

\bibitem[{Turchi et~al.(2017)Turchi, Negri, Farajian, and
  Federico}]{Turchi2017}
Marco Turchi, Matteo Negri, M.~Amin Farajian, and Marcello Federico. 2017.
\newblock \href {https://doi.org/10.1515/pralin-2017-0023} {{Continuous
  Learning from Human Post-Edits for Neural Machine Translation}}.
\newblock \emph{The Prague Bulletin of Mathematical Linguistics},
  108(1):233--244.

\bibitem[{Vaswani et~al.(2017)Vaswani, Shazeer, Parmar, Uszkoreit, Jones,
  Gomez, Kaiser, and Polosukhin}]{Vaswani2017}
Ashish Vaswani, Noam Shazeer, Niki Parmar, Jakob Uszkoreit, Llion Jones,
  Aidan~N. Gomez, {\L{}}ukasz Kaiser, and Illia Polosukhin. 2017.
\newblock \href {http://arxiv.org/abs/1706.03762} {{Attention is all you
  need}}.
\newblock In \emph{31st Conference on Neural Information Processing Systems
  (NIPS 2017)}, pages 5999--6009.

\bibitem[{Xia et~al.(2019)Xia, Tan, Tian, Gao, Chen, Fan, Gong, Leng, Luo,
  Wang, Wu, Zhu, Qin, and Liu}]{Xia2019}
Yingce Xia, Xu~Tan, Fei Tian, Fei Gao, Weicong Chen, Yang Fan, Linyuan Gong,
  Yichong Leng, Renqian Luo, Yiren Wang, Lijun Wu, Jinhua Zhu, Tao Qin, and
  Tie-Yan Liu. 2019.
\newblock \href {https://doi.org/10.18653/v1/W19-5348} {{Microsoft Research
  Asia's Systems for WMT19}}.
\newblock In \emph{Proceedings of the Fourth Conference on Machine Translation
  (Volume 2: Shared Task Papers, Day 1)}, pages 424--433, Stroudsburg, PA, USA.
  Association for Computational Linguistics.

\end{thebibliography}
%\bibliography{src/0refs}

%%%%%%%%%%%%%%%%%%%%%%%%%%%%%%%%%%%%%%%%%%%%%%%%%%%%%%%%%%%%%%%%%%%%%%%%%%%%%%%%
% APPENDIX
\appendix

\section{Weight evolution (all language pairs)}
\label{sec:appendixA}

Here we present the weight evolution per \acs{MT} system for all the combinations of language pairs, learning algorithms (\acs{EWAF} or \acs{EXP3}), and loss functions ({\tt human-zero}, {\tt human-avg}, or {\tt human-comet}, when applicable) -- except for those combinations that are already part of the main document. 

\subsection{English $\rightarrow$ German ({\tt en-de})}

\begin{figure}[!h]
\centering
\includegraphics[width=1\linewidth]{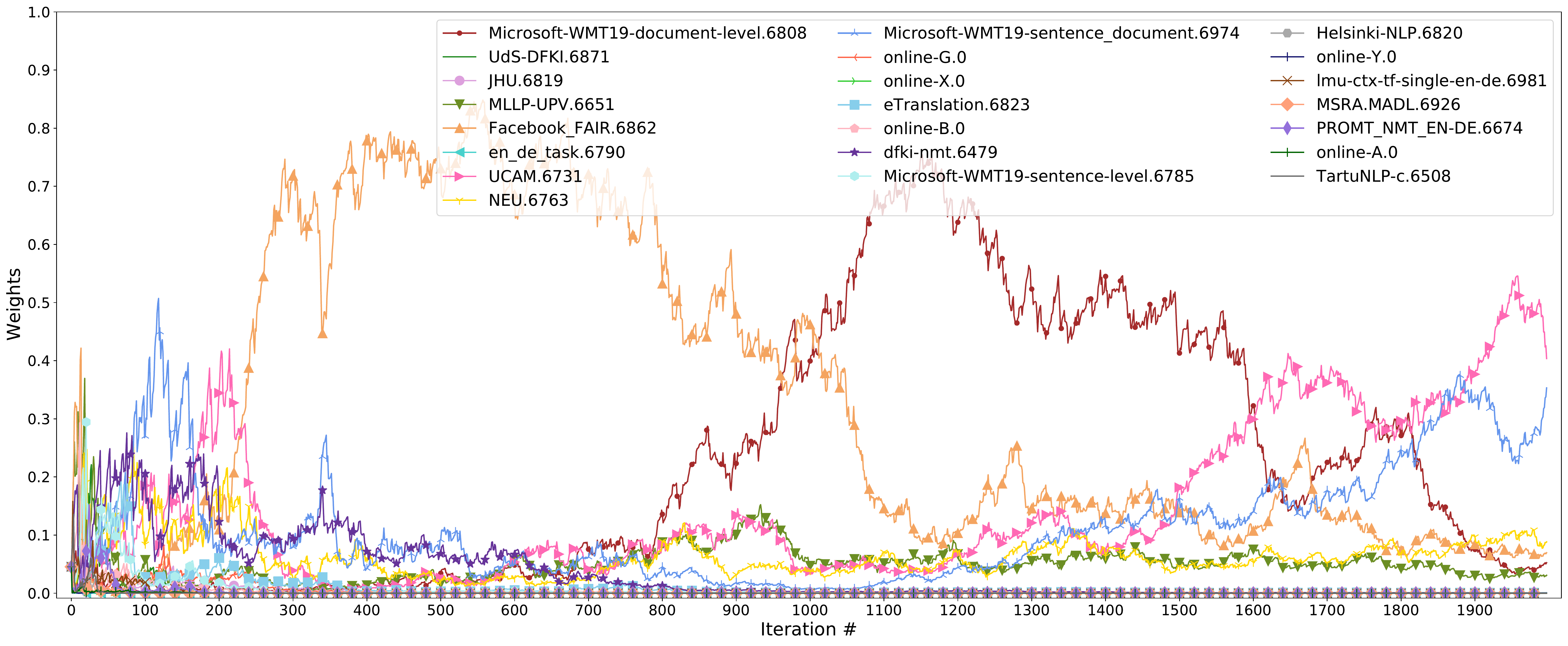}
\caption{\acs{EWAF} with {\tt human-zero} loss. Recall that, for this language pair, the official top 3 systems were Facebook-FAIR, Microsoft-sent-doc, and Microsoft-doc-level.}
%\caption{Weight evolution per \acs{MT} system when using \acs{EWAF} and {\tt human-zero} loss function ({\tt en-de}).}
\label{fig:weightsEnDeEWAFhuman}
\end{figure}

\begin{figure}[!h]
\centering
\includegraphics[width=1\linewidth]{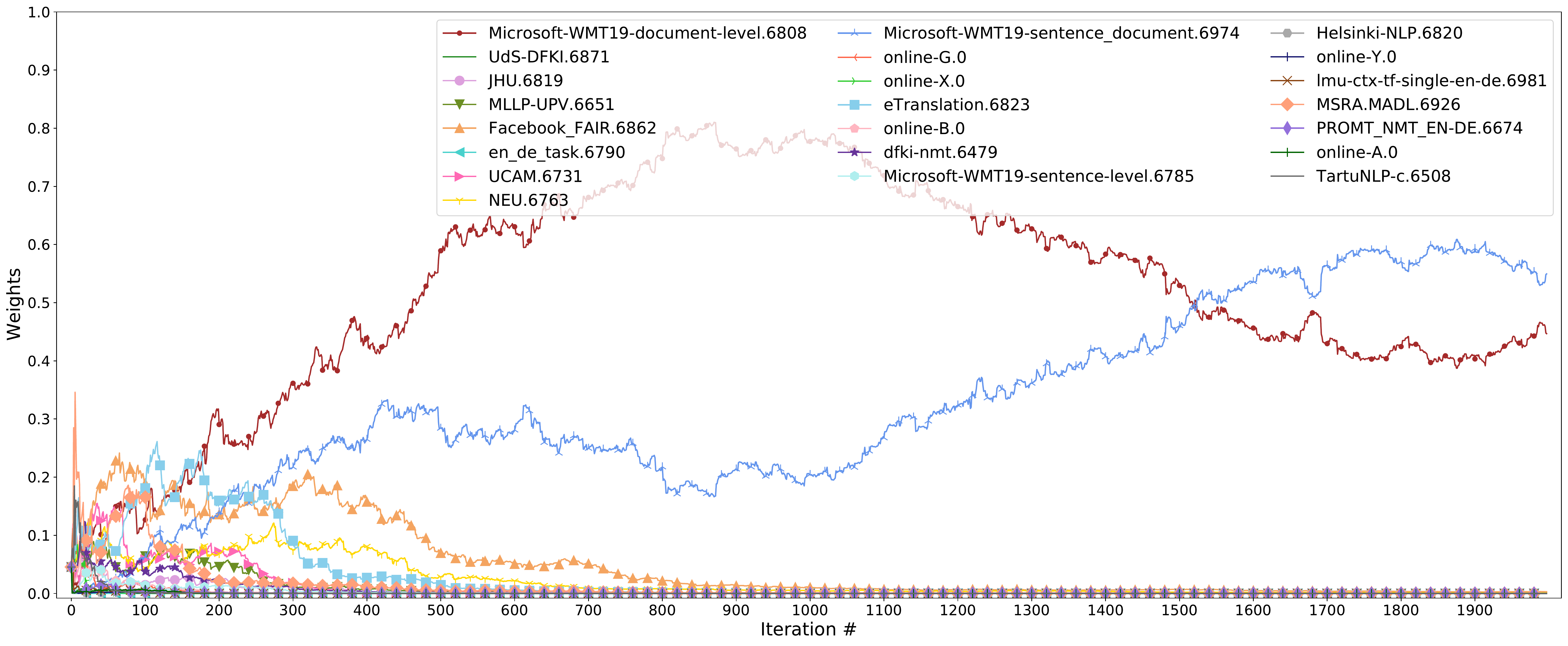}
\caption{\acs{EWAF} with {\tt human-avg} loss.}
% \caption{Weight evolution per \acs{MT} system when using \acs{EWAF} and {\tt human-avg} loss function ({\tt en-de}).}
\label{fig:weightsEnDeEWAFhumanavg}
\end{figure}

\begin{figure}[!h]
\centering
\includegraphics[width=1\linewidth]{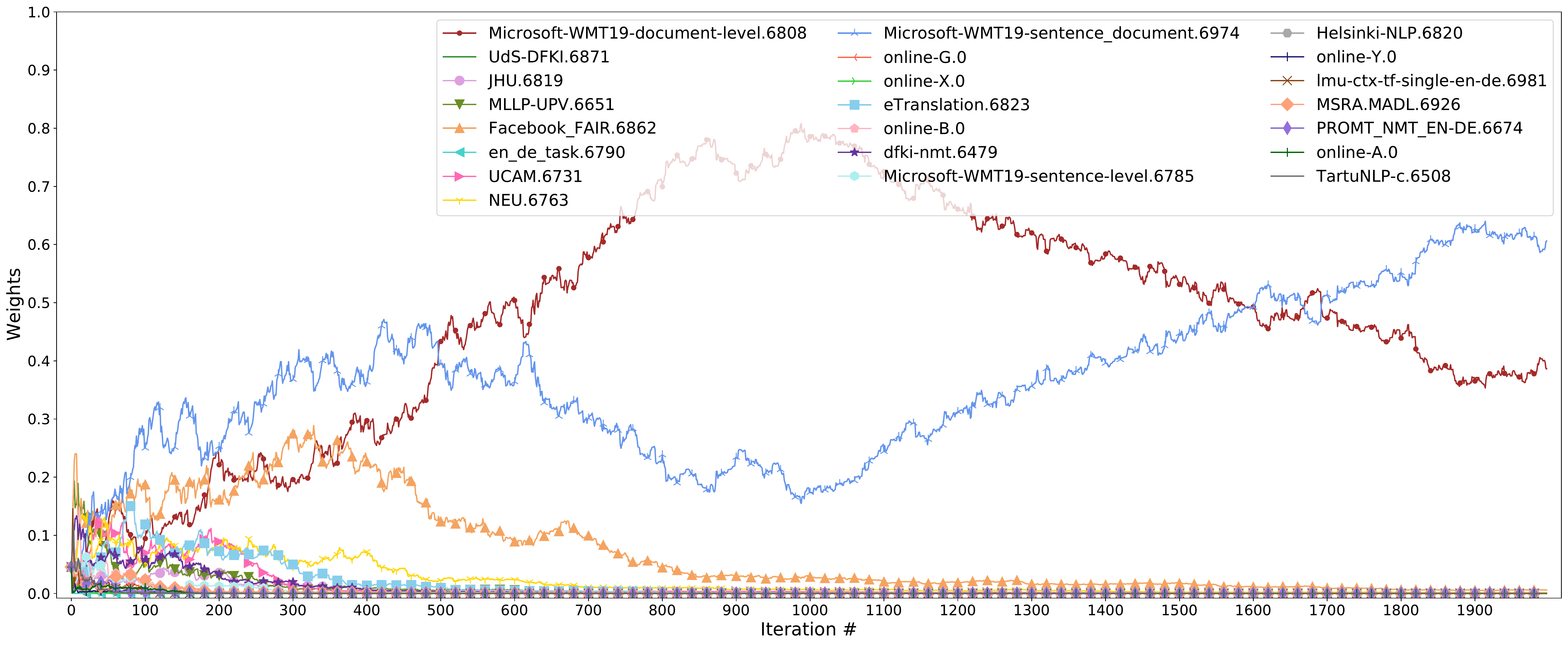}
\caption{\acs{EWAF} with {\tt human-comet} loss.}
% \caption{Weight evolution per \acs{MT} system when using \acs{EWAF} and {\tt human-comet} loss function ({\tt en-de}).}
\label{fig:weightsEnDeEWAFhumancomet}
\end{figure}

\begin{figure}[!h]
\centering
\includegraphics[width=1\linewidth]{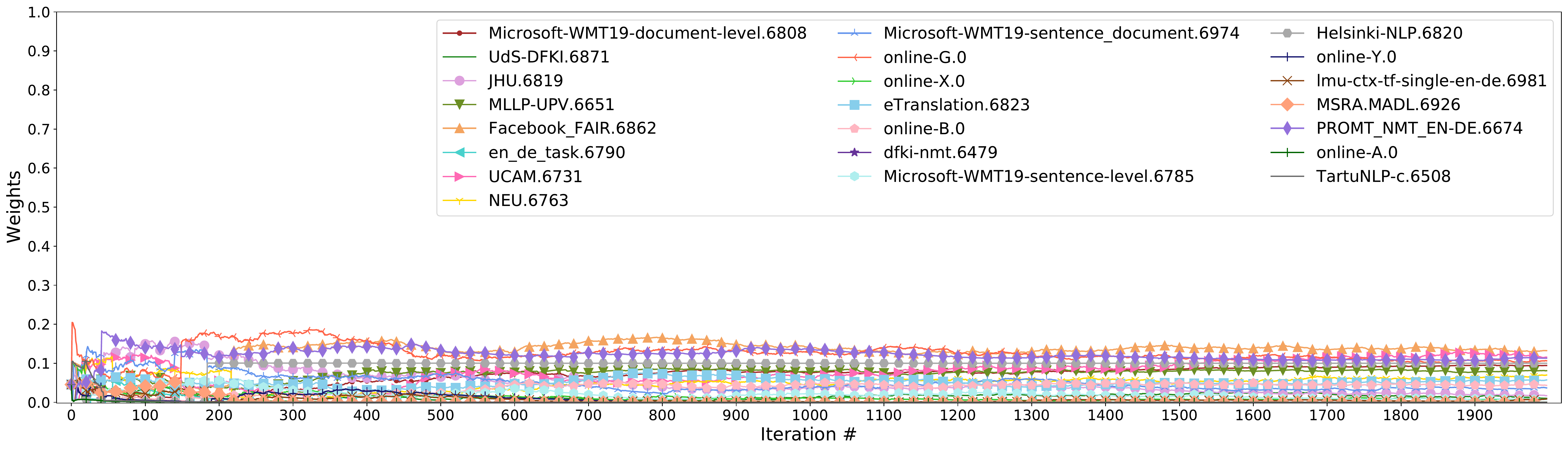}
\caption{\acs{EXP3} with {\tt human-zero} loss.}
% \caption{Weight evolution per \acs{MT} system when using \acs{EXP3} and {\tt human-zero} loss function ({\tt en-de}).}
\label{fig:weightsEnDeEXP3human}
\end{figure}

\begin{figure}[!h]
\centering
\includegraphics[width=1\linewidth]{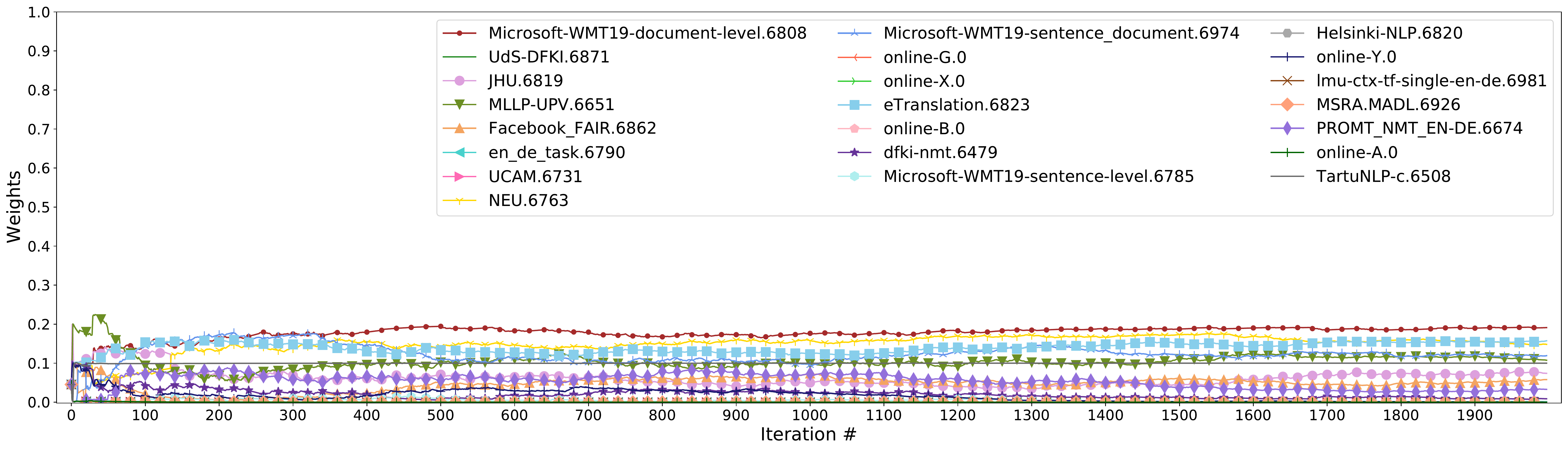}
\caption{\acs{EXP3} with {\tt human-avg} loss.}
% \caption{Weight evolution per \acs{MT} system when using \acs{EXP3} and {\tt human-avg} loss function ({\tt en-de}).}
\label{fig:weightsEnDeEXP3humanavg}
\end{figure}

\begin{figure}[!h]
\centering
\includegraphics[width=1\linewidth]{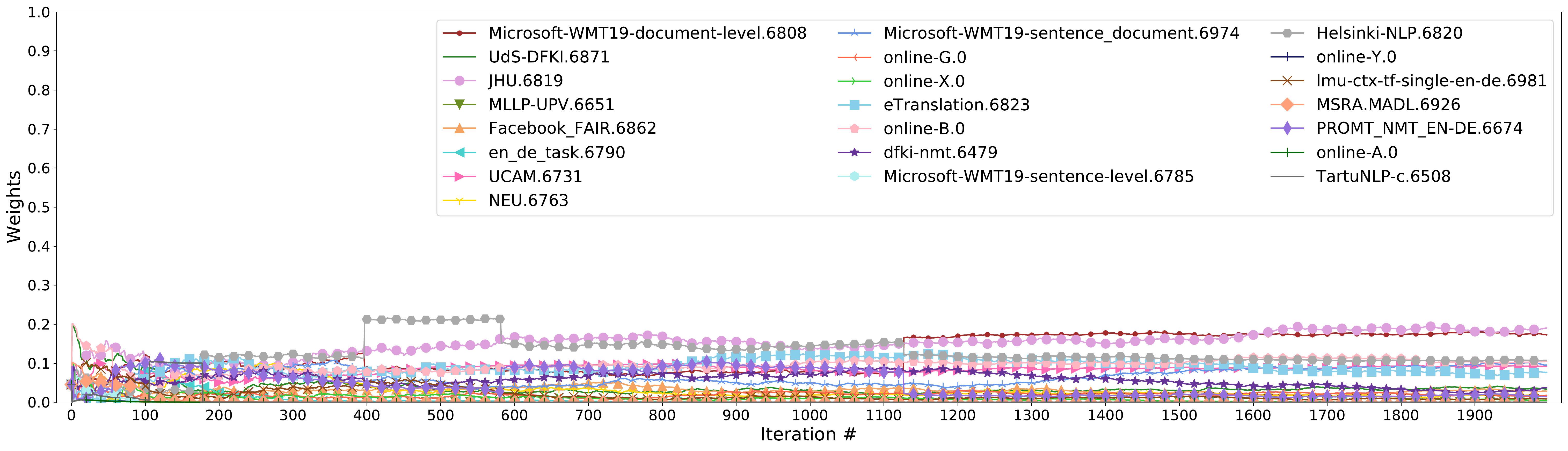}
\caption{\acs{EXP3} with {\tt human-comet} loss.}
% \caption{Weight evolution per \acs{MT} system when using \acs{EXP3} and {\tt human-comet} loss function ({\tt en-de}).}
\label{fig:weightsEnDeEXP3humancomet}
\end{figure}

\newpage
\subsection{French $\rightarrow$ German ({\tt fr-de})}

\begin{figure}[!h]
\centering
\includegraphics[width=1\linewidth]{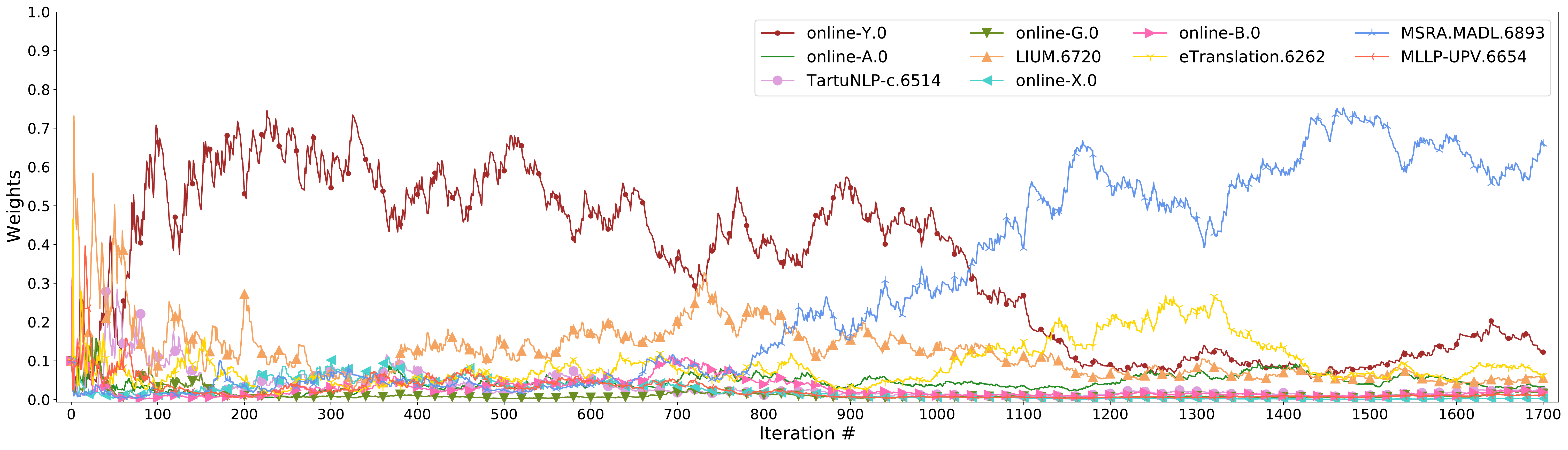}
\caption{\acs{EWAF} with {\tt human-zero} loss. Recall that, for this language pair, the official top 3 systems were MSRA-MADL, eTranslation, and LIUM.}
% \caption{Weight evolution per \acs{MT} system when using \acs{EWAF} and {\tt human-zero} loss function ({\tt fr-de}).}
\label{fig:weightsFrDeEWAFhuman}
\end{figure}

\begin{figure}[!h]
\centering
\includegraphics[width=1\linewidth]{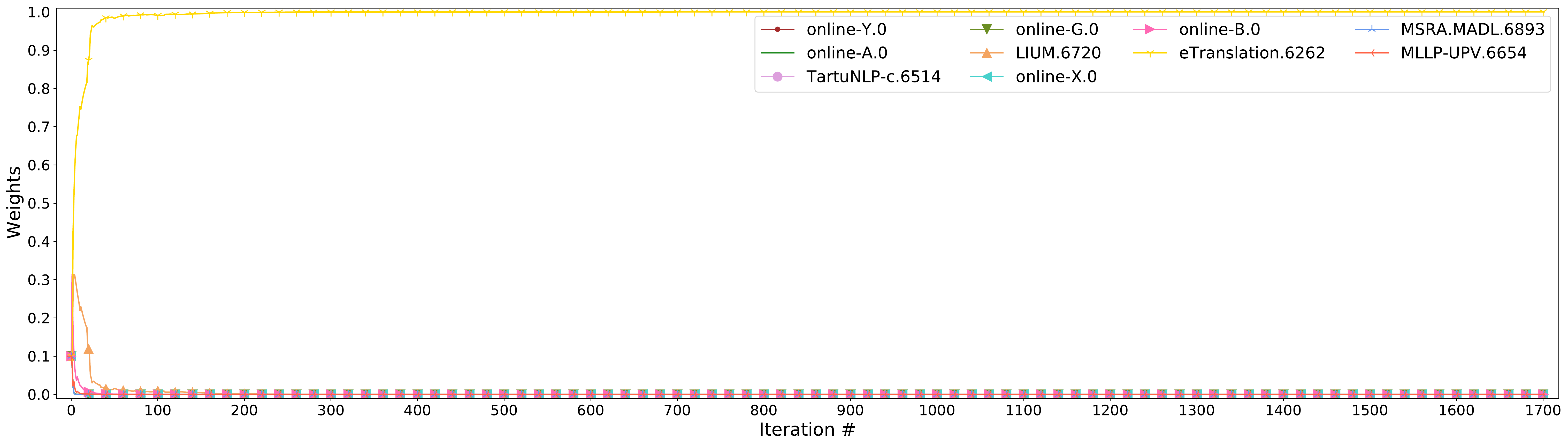}
\caption{\acs{EWAF} with {\tt human-avg} loss.}
% \caption{Weight evolution per \acs{MT} system when using \acs{EWAF} and {\tt human-avg} loss function ({\tt fr-de}).}
\label{fig:weightsFrDeEWAFhumanavg}
\end{figure}

\begin{figure}[!h]
\centering
\includegraphics[width=1\linewidth]{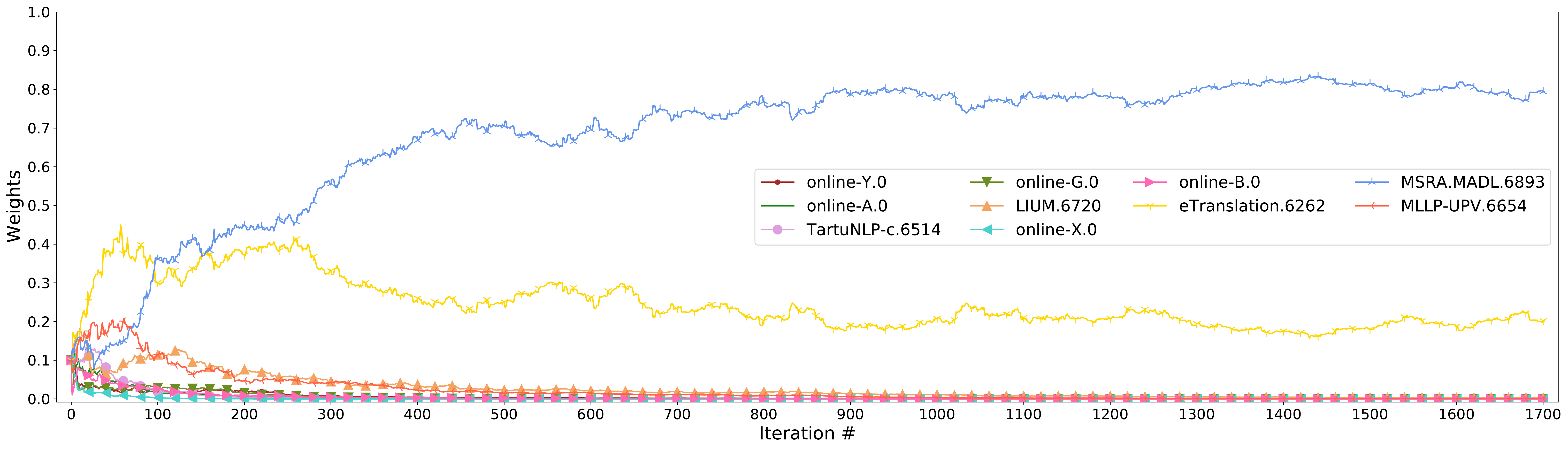}
\caption{\acs{EWAF} with {\tt human-comet} loss.}
% \caption{Weight evolution per \acs{MT} system when using \acs{EWAF} and {\tt human-comet} loss function ({\tt fr-de}).}
\label{fig:weightsFrDeEWAFhumancomet}
\end{figure}

\begin{figure}[!h]
\centering
\includegraphics[width=1\linewidth]{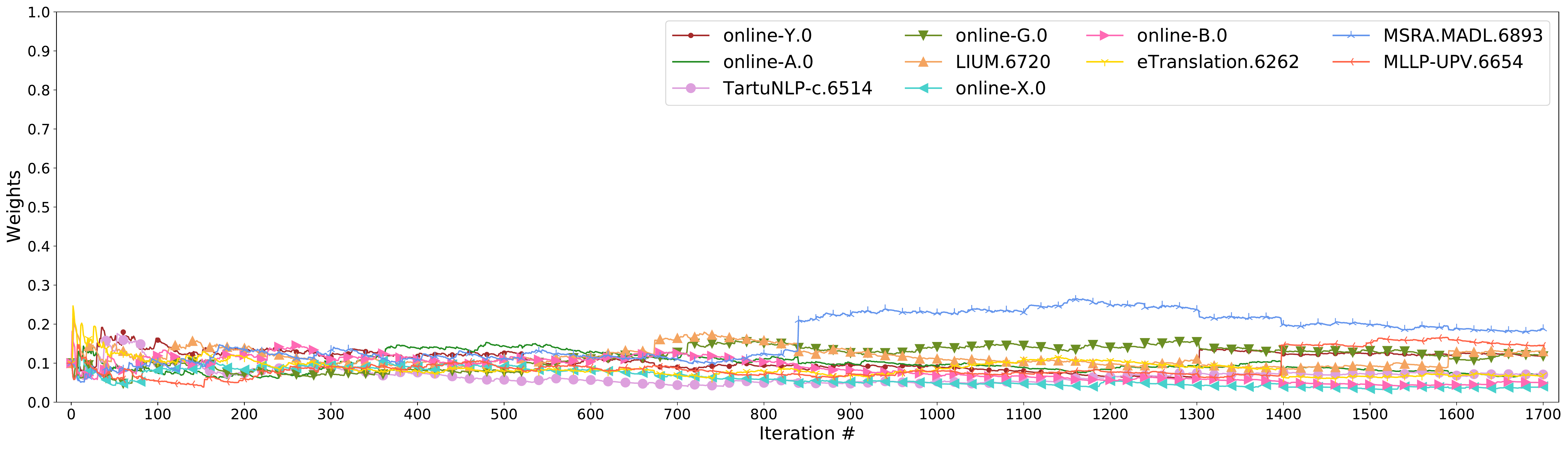}
\caption{\acs{EXP3} with {\tt human-zero} loss.}
% \caption{Weight evolution per \acs{MT} system when using \acs{EXP3} and {\tt human-zero} loss function ({\tt fr-de}).}
\label{fig:weightsFrDeEXP3human}
\end{figure}

\begin{figure}[!h]
\centering
\includegraphics[width=1\linewidth]{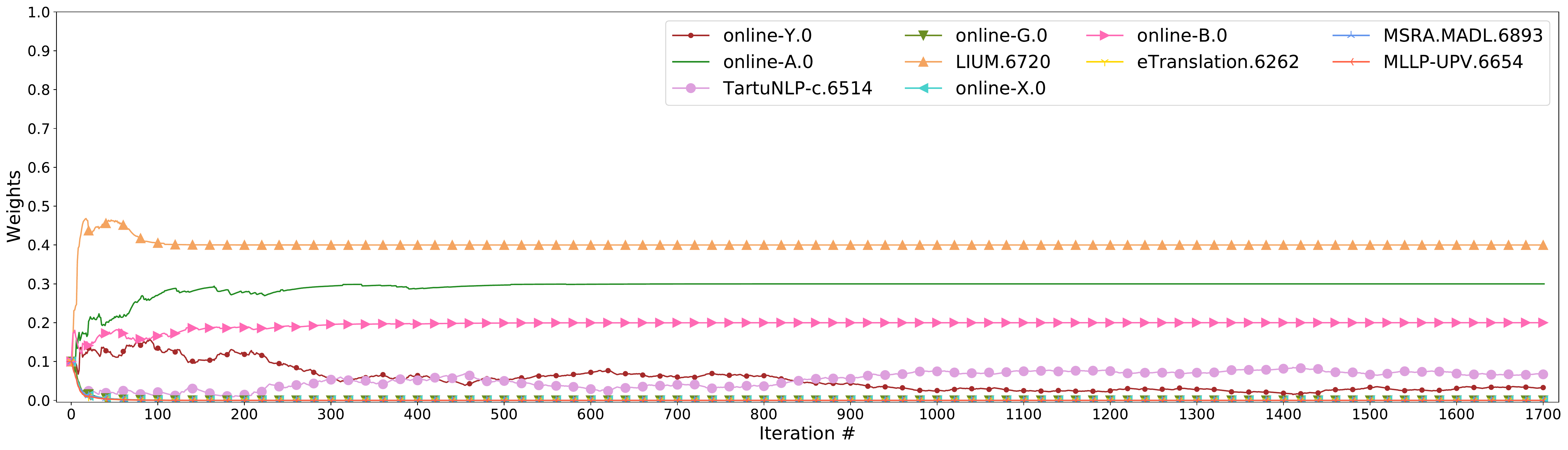}
\caption{\acs{EXP3} with {\tt human-avg} loss.}
% \caption{Weight evolution per \acs{MT} system when using \acs{EXP3} and {\tt human-avg} loss function ({\tt fr-de}).}
\label{fig:weightsFrDeEXP3humanavg}
\end{figure}

\begin{figure}[!h]
\centering
\includegraphics[width=1\linewidth]{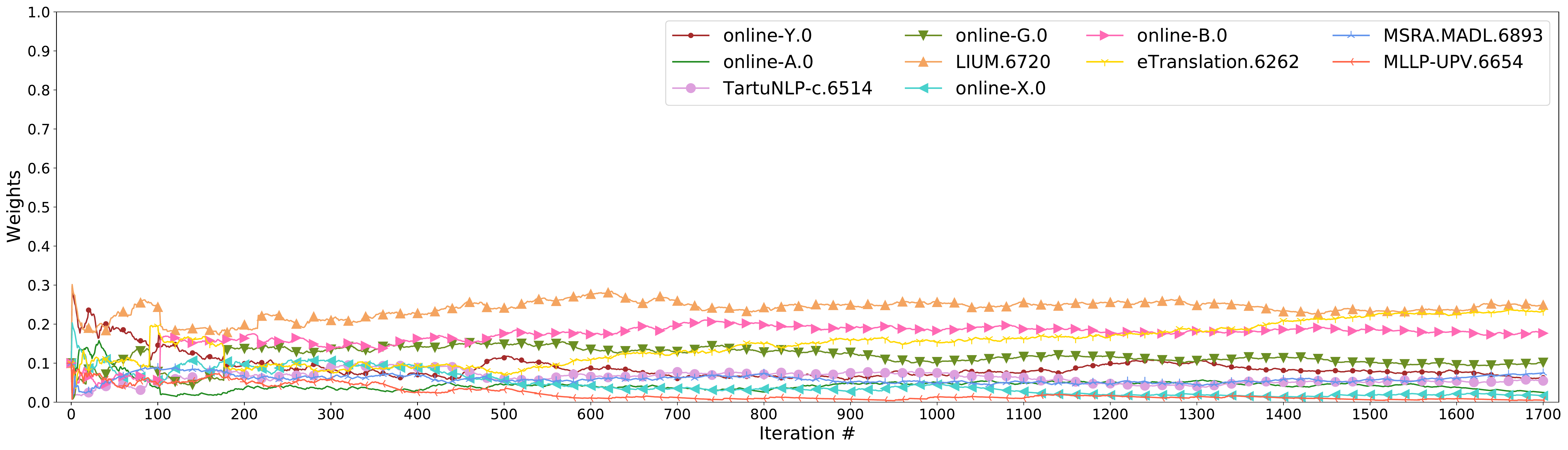}
\caption{\acs{EXP3} with {\tt human-comet} loss.}
% \caption{Weight evolution per \acs{MT} system when using \acs{EXP3} and {\tt human-comet} loss function ({\tt fr-de}).}
\label{fig:weightsFrDeEXP3humancomet}
\end{figure}

\newpage
\subsection{German $\rightarrow$ Czech ({\tt de-cs})}

% \begin{figure*}[!h]
% \centering
% \includegraphics[width=1\linewidth]{weightsDeCsEWAFhuman}
% \caption{Weight evolution per \acs{MT} system when using \acs{EWAF} and {\tt human-zero} loss function ({\tt de-cs}).}
% \label{fig:weightsEWAFhuman}
% \end{figure*}

% \begin{figure*}[!h]
% \centering
% \includegraphics[width=1\linewidth]{weightsDeCsEWAFhumanavg}
% \caption{Weight evolution per \acs{MT} system when using \acs{EWAF} and {\tt human-avg} loss function ({\tt de-cs}).}
% \label{fig:weightsEWAFhumanavg}
% \end{figure*}

% \begin{figure*}[!h]
% \centering
% \includegraphics[width=1\linewidth]{weightsDeCsEWAFhumancomet}
% \caption{Weight evolution per \acs{MT} system when using \acs{EWAF} and {\tt human-comet} loss function ({\tt de-cs}).}
% \label{fig:weightsEWAFhumancomet}
% \end{figure*}

% \begin{figure*}[!h]
% \centering
% \includegraphics[width=1\linewidth]{weightsDeCsEXP3human}
% \caption{Weight evolution per \acs{MT} system when using \acs{EXP3} and {\tt human-zero} loss function ({\tt de-cs}).}
% \label{fig:weightsEXP3human}
% \end{figure*}

\begin{figure}[!h]
\centering
\includegraphics[width=1\linewidth]{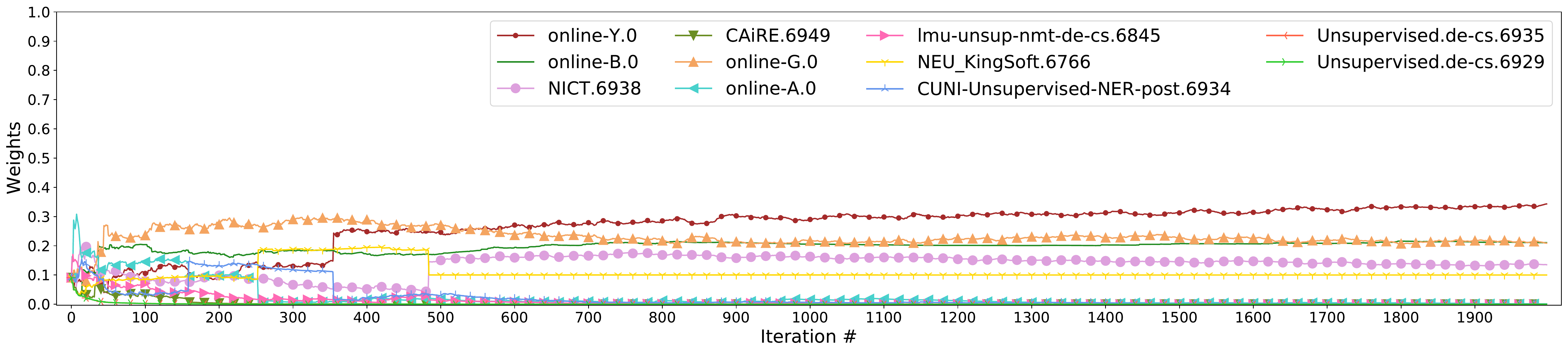}
\caption{\acs{EXP3} with {\tt human-avg} loss. Recall that, for this language pair, the official top 3 systems were online-Y, online-B, and NICT.}
% \caption{Weight evolution per \acs{MT} system when using \acs{EXP3} and {\tt human-avg} loss function ({\tt de-cs}).}
\label{fig:weightsDeCsEXP3humanavg}
\end{figure}

\begin{figure}[!h]
\centering
\includegraphics[width=1\linewidth]{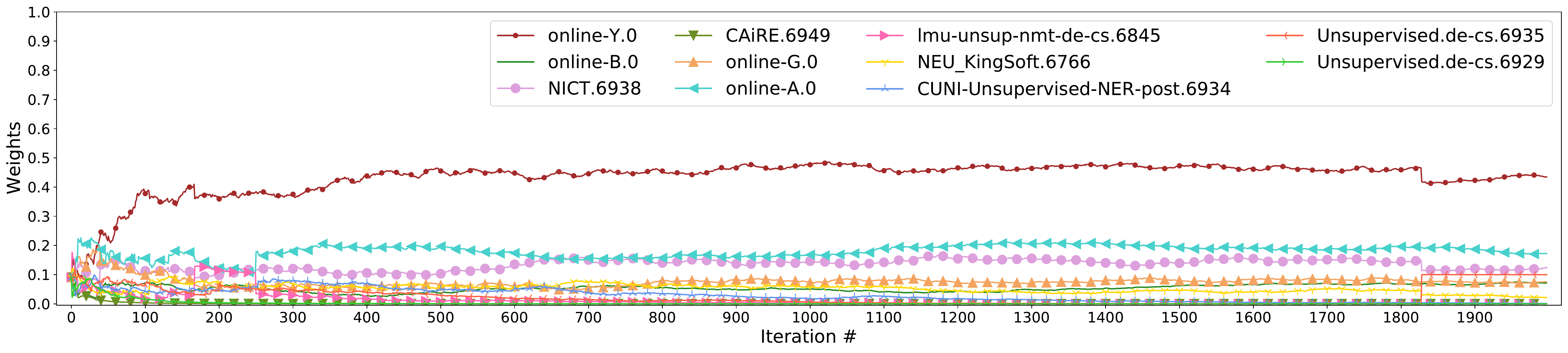}
\caption{\acs{EXP3} with {\tt human-comet} loss.}
% \caption{Weight evolution per \acs{MT} system when using \acs{EXP3} and {\tt human-comet} loss function ({\tt de-cs}).}
\label{fig:weightsDeCsEXP3humancomet}
\end{figure}

%\newpage
\subsection{Gujarati $\rightarrow$ English ({\tt gu-en})}

\begin{figure}[!h]
\centering
\includegraphics[width=1\linewidth]{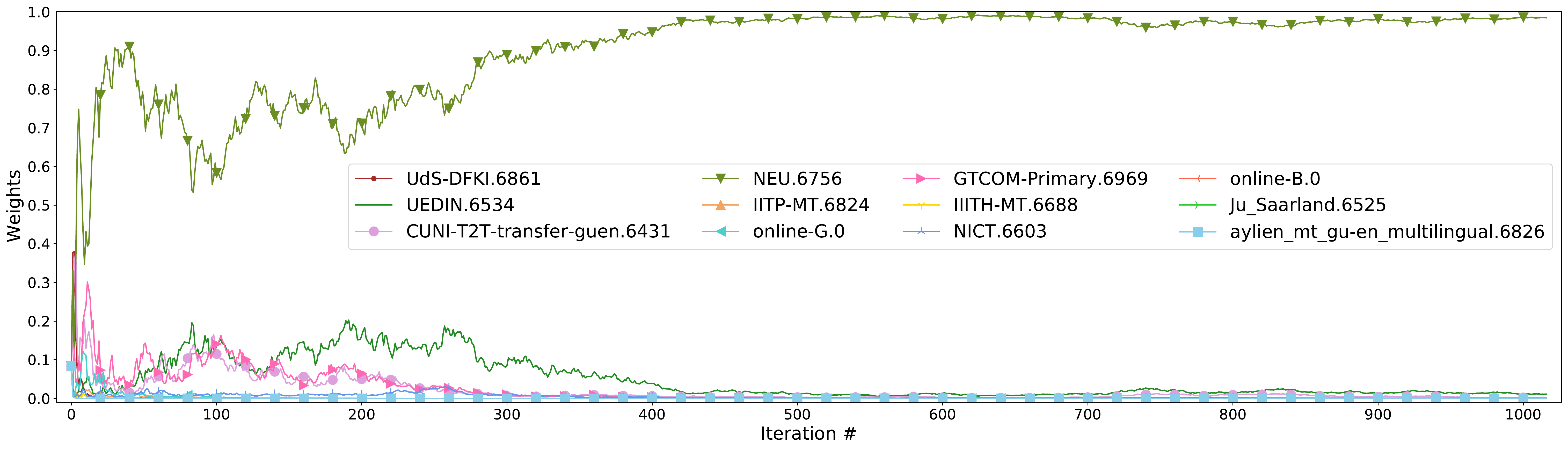}
\caption{\acs{EWAF} with {\tt human-zero} loss. Recall that, for this language pair, the official top 3 systems were NEU, UEDIN, and GTCOM-Primary.}
% \caption{Weight evolution per \acs{MT} system when using \acs{EWAF} and {\tt human-zero} loss function ({\tt gu-en}).}
\label{fig:weightsGuEnEWAFhuman}
\end{figure}

\begin{figure}[!h]
\centering
\includegraphics[width=1\linewidth]{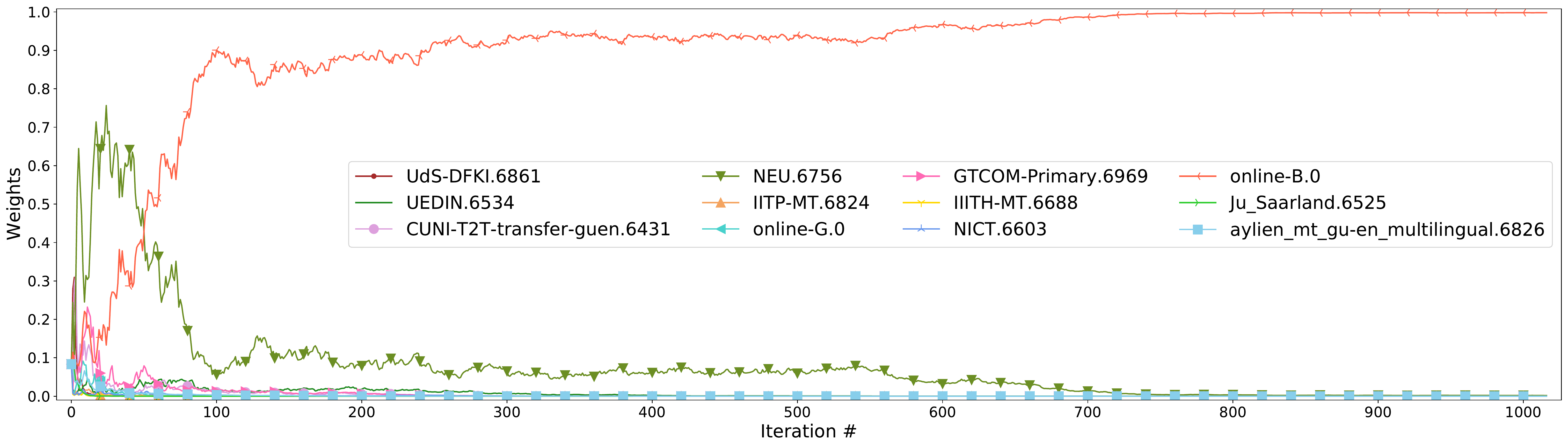}
\caption{\acs{EWAF} with {\tt human-comet} loss.}
% \caption{Weight evolution per \acs{MT} system when using \acs{EWAF} and {\tt human-comet} loss function ({\tt gu-en}).}
\label{fig:weightsGuEnEWAFhumancomet}
\end{figure}

\begin{figure}[!h]
\centering
\includegraphics[width=1\linewidth]{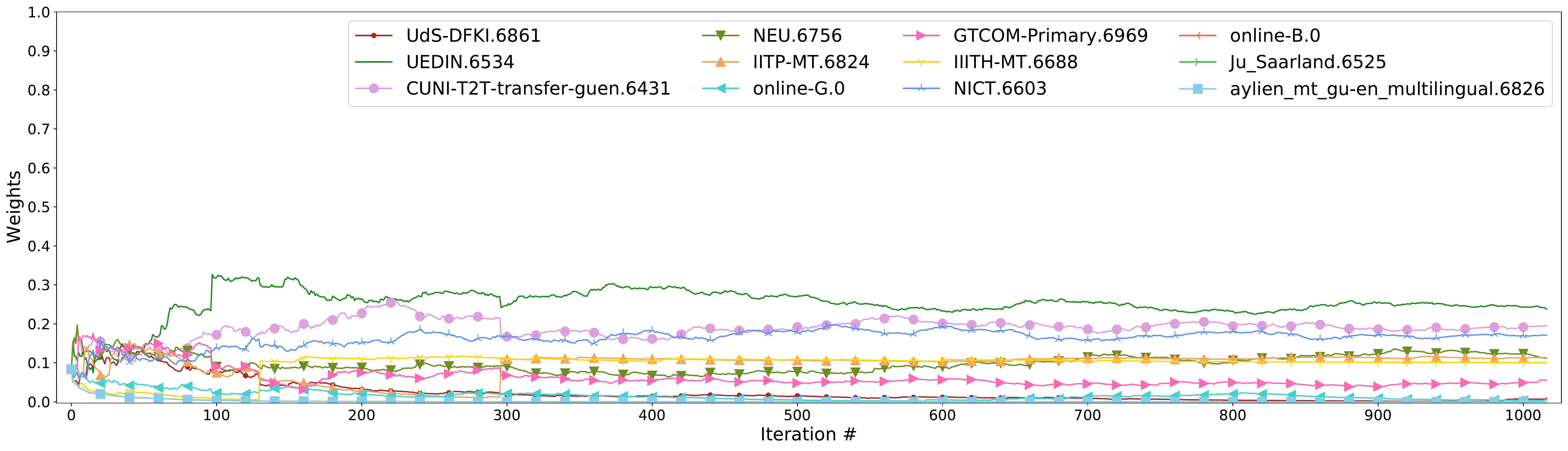}
\caption{\acs{EXP3} with {\tt human-zero} loss.}
% \caption{Weight evolution per \acs{MT} system when using \acs{EXP3} and {\tt human-zero} loss function ({\tt gu-en}).}
\label{fig:weightsGuEnEXP3human}
\end{figure}

\begin{figure}[!h]
\centering
\includegraphics[width=1\linewidth]{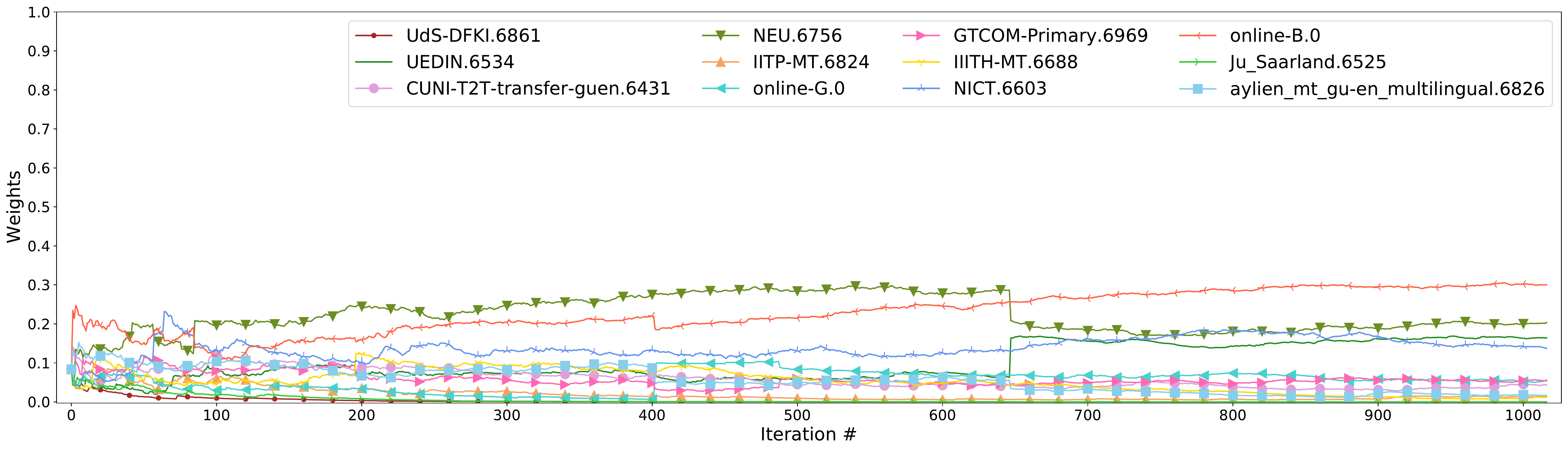}
\caption{\acs{EXP3} with {\tt human-comet} loss.}
% \caption{Weight evolution per \acs{MT} system when using \acs{EXP3} and {\tt human-comet} loss function ({\tt gu-en}).}
\label{fig:weightsGuEnEXP3humancomet}
\end{figure}

\newpage
\subsection{Lithuanian $\rightarrow$ English ({\tt lt-en})}

\begin{figure}[!h]
\centering
\includegraphics[width=1\linewidth]{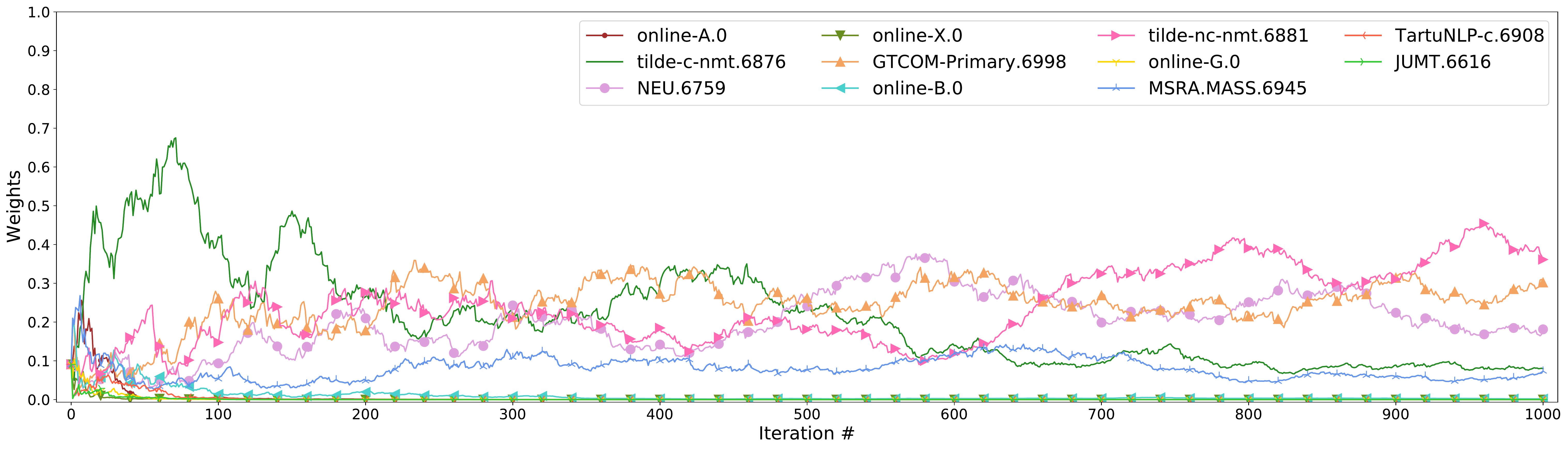}
\caption{\acs{EWAF} with {\tt human-zero} loss. Recall that, for this language pair, the official top 3 systems were GTCOM-Primary, tilde-nc-nmt, and NEU.}
% \caption{Weight evolution per \acs{MT} system when using \acs{EWAF} and {\tt human-zero} loss function ({\tt lt-en}).}
\label{fig:weightsLtEnEWAFhuman}
\end{figure}

\begin{figure}[!h]
\centering
\includegraphics[width=1\linewidth]{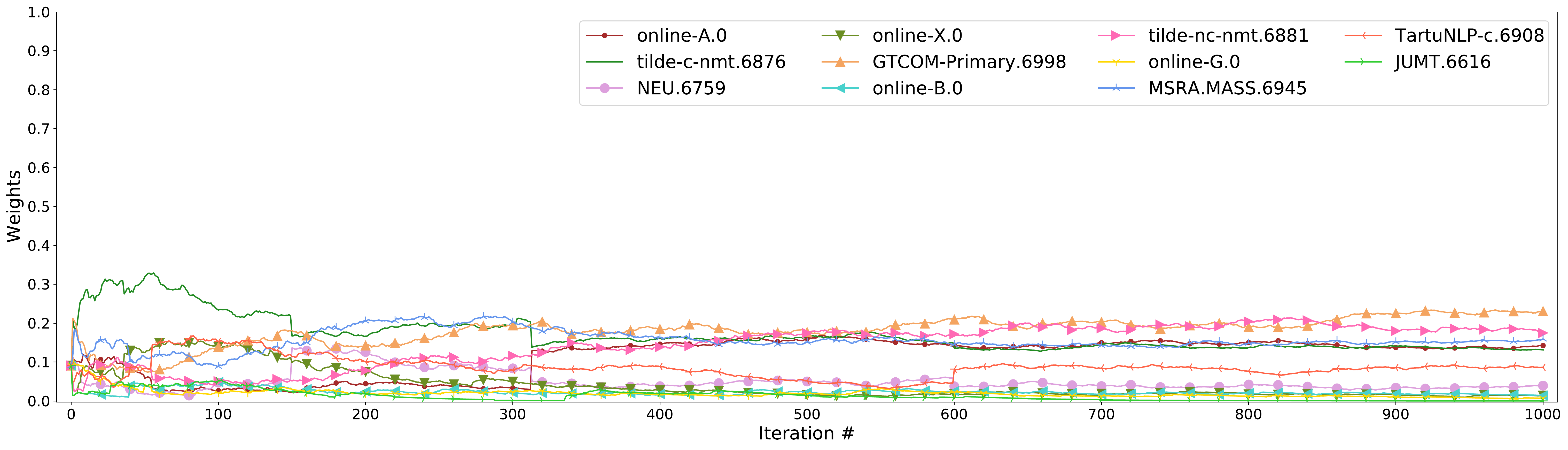}
\caption{\acs{EXP3} and {\tt human-zero} loss.}
% \caption{Weight evolution per \acs{MT} system when using \acs{EXP3} and {\tt human-zero} loss function ({\tt lt-en}).}
\label{fig:weightsLtEnEXP3human}
\end{figure}

\end{document}